%% file: eccv2020submission.tex
\documentclass[runningheads]{llncs}
\usepackage{graphicx}

\usepackage{booktabs}
\usepackage{tikz}
\usepackage{comment}
\usepackage{amsmath,amssymb} 
\usepackage{color}
\usepackage{multicol}
\usepackage{multirow}
\usepackage{subcaption}
\usepackage{hyperref}
\input{macros}

\begin{document}
\pagestyle{headings}
\mainmatter
\def\ECCVSubNumber{100}  

\title{SHARP 2020: The 1st Shape Recovery from Partial Textured 3D Scans Challenge Results } 

\titlerunning{ECCV-20 submission ID \ECCVSubNumber} 
\authorrunning{ECCV-20 submission ID \ECCVSubNumber} 
\author{Anonymous ECCV submission}
\institute{Paper ID \ECCVSubNumber}

\titlerunning{SHARP 2020: Challenge Results}
%
\author{
  Alexandre Saint\inst{1} \and
  Anis Kacem\inst{1} \and
  Kseniya Cherenkova\inst{1,2} \and
  Konstantinos Papadopoulos\inst{1} \and
  Julian Chibane\inst{3} \and
  Gerard Pons-Moll\inst{3} \and
  Gleb Gusev\inst{2} \and
  David Fofi\inst{4} \and
  Djamila Aouada\inst{1} \and
  Bj\"orn Ottersten\inst{1}
}
\authorrunning{A. Saint et al.}
%
\institute{
  SnT, University of Luxembourg \\
  \email{firstname.lastname@uni.lu} \and
  Artec 3D, Luxembourg \\
  \email{\{kcherenkova,gleb\}@artec-group.com} \and
  Max Planck Institute for Informatics\\
  \email{\{jchibane,gpons\}@mpi-inf.mpg.de}\and
  University of Burgundy\\
  \email{david.fofi@u-bourgogne.fr}
}
\maketitle


\input{abstract}
\input{introduction}
\input{datasets}
\input{evaluation}
\input{results}
\input{conclusion}
\input{acknowledgements}

\clearpage
%
%
\bibliographystyle{splncs04}
\bibliography{egbib}
\end{document}

%% file: macros.tex
\usepackage{xspace}
\makeatletter
\DeclareRobustCommand\onedot{\futurelet\@let@token\@onedot}
\def\@onedot{\ifx\@let@token.\else.\null\fi\xspace}
\def\eg{\emph{e.g}\onedot}

\def\ie{\emph{i.e}\onedot}

%% file: abstract.tex
\begin{abstract}
The \emph{SHApe Recovery from Partial textured 3D scans} challenge, SHARP 2020,
is the first edition of a challenge fostering and benchmarking methods for
recovering complete textured 3D scans from raw incomplete data.
SHARP 2020 is organised as a workshop in conjunction with ECCV 2020.
There are two complementary challenges,
the first one on 3D human scans,
and the second one on generic objects.
Challenge 1 is further split into two tracks,
focusing, first, on large body and clothing regions,
and, second, on fine body details.
A novel evaluation metric is proposed to quantify jointly
the shape reconstruction,
the texture reconstruction
and the amount of completed data.
Additionally, two unique datasets of 3D scans are proposed, to provide raw
ground-truth data for the benchmarks.
The datasets are released to the scientific community.
Moreover, an accompanying custom library of software routines is also released
to the scientific community.
It allows for processing 3D scans,
generating partial data
and performing the evaluation.
Results of the competition, analysed in comparison to baselines,
show the validity of the proposed evaluation metrics,
and highlight the challenging aspects of the task and of the datasets.
Details on the SHARP 2020 challenge can be found at
\url{https://cvi2.uni.lu/sharp2020/}.
\end{abstract}

%% file: introduction.tex
\section{Introduction}
\label{sec:introduction}

Representing the physical world in 3D, including shape and
colour, is key for industrial and research purposes~\cite{%
  anguelov2005scape,%
  yang2014semantic,%
  hasler2009statistical,%
  pishchulin2017building,%
  bogo2014faust,%
  xu2018multilevel,%
  bronstein2008numerical,%
  robinette1999caesar%
}.
It includes, for example, areas from virtual reality to heritage conservation,
or from medical treatment to fitness, entertainment and fashion.
3D scanning allows to digitise the physical world, \eg objects and humans.
Acquired 3D scans vary in quality depending
on the scanning system used,
the properties of the target and the environment.
For example, a high-end photogrammetric scanning system with a fixed camera
array might capture high-quality data at a high frame rate
but might be bulky, have a fixed structure, suffer from occlusion
and limited in scanning volume.
Other high-end systems such as hand-held devices might produce accurate
results, while easily transported and easily oriented to limit occlusion,
but cannot handle movable targets and are time-consuming.
On the other hand, low-end scanning systems might be flexible and easy to manipulate but produce
low-quality scans.
Limiting factors due to the target or the environment include
varying levels of details (\eg finer anatomical parts),
occlusion, non-rigidity, movement, and optical properties
(\eg fabric, material, hair and reflection).
Moreover, for time-consuming acquisition systems or moving targets, it might
be desirable or only possible to capture partial data.
In this work, defective and/or partial acquisitions, are both viewed as
data with missing information that must be completed.

The SHARP 2020 challenge for
\emph{SHApe Recovery from Partial textured 3D scans}
is proposed to foster research and provide a benchmark on 3D shape and texture
completion from partial 3D scan data.
First, two new unique datasets of 3D textured scans are proposed to serve
as reference data.
These datasets contain thousands of scans of humans and generic objects with varied
identities, clothing, colours, shapes and categories.
One challenge is proposed per dataset,
challenge 1 focusing on human scans
and challenge 2 on object scans.
Second, partial scans are generated synthetically, but randomly, to simulate
a general pattern of partial data acquisition while still having access
to ground truth data.
Third, specific evaluation metrics are proposed to quantitatively measure
the quality of the shape and texture reconstructions,
and the amount of completed data.
Fourth, reusable software libraries developed for the challenge are also
made available.
These contain routines
to process 3D scans,
to generate partial data,
and to evaluate and analyse the submissions on the proposed benchmark.
This paper summaries the SHARP challenge
with a presentation of the proposed datasets, benchmark and evaluation method,
as well as the results of the submitted methods
and an analysis of the results.
SHARP 2020 is the first edition of the challenge,
held in conjunction with the 16th European Conference on Computer Vision (ECCV).

In the following,
Section~\ref{sec:datasets} describes the challenge and the proposed datasets.
Section~\ref{sec:evaluation} describes the proposed evaluation protocol and,
in Section~\ref{sec:results}, an extensive analysis on the results is presented.
To conclude, the results and outcomes of the challenge are discussed in
Section~\ref{sec:discussion}.

%% file: datasets.tex
\section{Challenges and Datasets}
\label{sec:datasets}

The SHARP challenge is split into two separate challenges:
challenge 1 focuses on human scans,
and challenge 2 focuses on generic object scans.
Two corresponding datasets are introduced,
3DBodyTex.v2 for human scans,
and 3DObjectTex for generic object scans.
Table~\ref{tab:datasets} describes the datasets
with figures on the number of samples for different subsets,
including the splits used in the challenges
and categories of body pose and clothing type for 3DBodyTex.v2.

\textbf{3DBodyTex.v2} is an extension of 3DBodyTex proposed by
Saint~et~al.~\cite{saint20183dbodytex}.
See sample scans in Fig.~\ref{fig:3dbodytex}.
It contains about 3000 static 3D human scans with high-resolution texture.
It features a large variety of poses and clothing types,
with about 500 different subjects.
Each subject is captured in about 3 poses.
Most subjects perform the corresponding poses in both
standard close-fitting clothing and arbitrary casual clothing.
The faces are anonymised, for privacy reasons, by blurring the shape and the
texture.

\textbf{3DObjectTex} is a subset of the
\emph{viewshape}~\cite{viewshape} repository.
See sample scans in Fig.~\ref{fig:3dobjecttex}.
It consists of about 1200 textured 3D scans of generic objects with
a large variation in categories and in physical dimensions.

Both datasets encode the scans as 3D triangle meshes.
The colour is encoded in a texture atlas with an independent and arbitrary
UV mapping~\cite{zayer2007linear} for each mesh.

\setlength{\tabcolsep}{6pt}
\begin{table}
\centering
\begin{tabular}{@{}lllccccll@{}}
\toprule
dataset & subset & & \multicolumn{4}{c}{samples} & challenge & track \\
\cmidrule(lr){4-7}
 & & & train & val & test & total & & \\
\midrule
\multirow{5}{*}{3DBodyTex.v2} & full & & 2094 & 454 & 451 & 2999 & 1 & 1 \\
\cmidrule(lr){2-7}
 & \multirow{2}{*}{clothing} & fitness & 844 & 178 & 182 & 1204 & 1 & 2 \\
 &          & casual & 1250 & 276 & 269 & 1795 & \,- & \,- \\
\cmidrule(lr){2-7}
 & \multirow{2}{*}{poses} & standard & 977 & 219 & 224 & 1420 & \,- & \,- \\
 &       & other & 1117 & 235 & 227 & 1579 & \,- & \,- \\
\midrule
 3DObjectTex & \,- & & 799 & 205 & 205 & 1209 & 2 & \,- \\
\bottomrule
\\
\end{tabular}
\caption{
  Contents of the datasets and categorical subsets,
  along with associated challenge and track.
  The \emph{standard poses} are the \emph{A} and \emph{U} rest poses.
  The \emph{other poses} are varied, from a predefined list or arbitrary.
  The \emph{casual} and \emph{fitness} clothing types are shown in
  Fig.~\ref{fig:3dbodytex}.
}
\label{tab:datasets}
\end{table}

\subsection{Challenge 1: Recovery of Human Body Scans}

Challenge 1 covers the reconstruction of human scans,
both with loose casual clothing and minimal close-fitting clothing.

\subsubsection{Track 1: Recovery of Large Regions.}

\begin{figure}
  \centering
  \includegraphics[width=\textwidth]{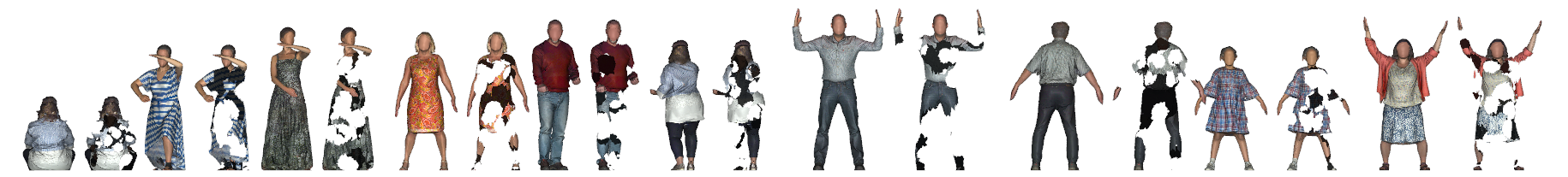}
  \includegraphics[width=\textwidth]{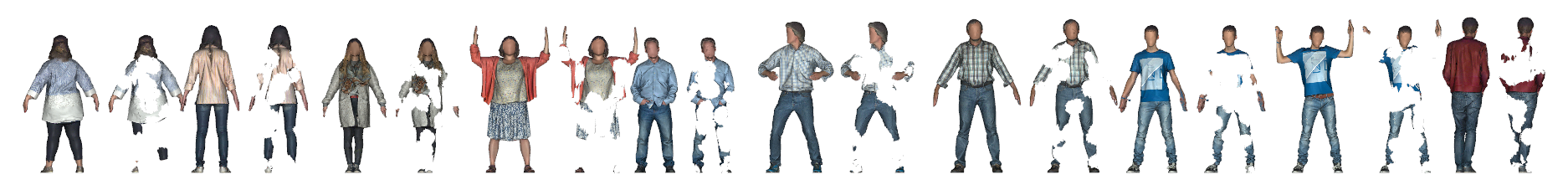}
  \includegraphics[width=\textwidth]{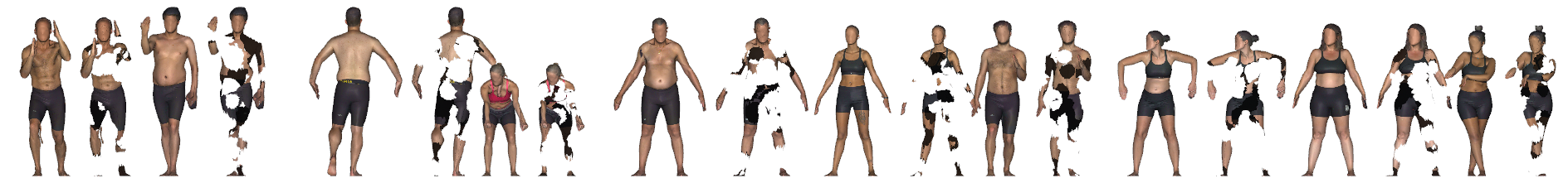}
  \caption{
    Samples of 3DBodyTex.v2 in casual (top rows) and fitness (bottom row)
    clothing.
    For each person,
    ground truth scan (left),
    sample synthetic partial scan (right).
  }
  \label{fig:3dbodytex}
\end{figure}

Track 1 focuses on the reconstruction of large regions of human scans,
excluding hands and head.
Fig.~\ref{fig:3dbodytex} shows samples of the data, with ground-truth
and partial scans.
Both shape and texture are considered.
These large regions are of relatively high quality in the raw reference data.
The fine details (hands and head) are of unreliable quality and thus ignored in
this track.
The full 3DBodyTex.v2 dataset is used (see Table~\ref{tab:datasets}).

\subsubsection{Track 2: Recovery of Fine Details.}

Track 2 focuses on fine body details not considered in Track 1,
\ie, hands, fingers, and nose.
The raw scans are not of reliable quality for these details
and the faces are not released due to privacy concerns.
Thus, the reference data is generated synthetically from a
body model~\cite{pavlakos2019expressive} and the scans of 3DBodyTex.v2
in \emph{fitness} clothing (see Table~\ref{tab:datasets}),
where the fine body details are not occluded, to capture to the real
distribution of rough poses and shapes.
The reference data is generated in two steps:
(1) fit a parametric body model to the scans to obtain the ground-truth data;
(2) simulate the scanning process in software to obtain a synthetic scan
(with simulated artefacts in the regions of interest, \ie hands, ears...)
from which the partial data is generated.
The fitting is performed with the approach of
Saint~et~al.~\cite{saint2017towards,saint20183dbodytex,saint2019bodyfitr}.
The texture is not considered in this setting as
the raw data does not contain texture of reliable quality
(and the parametric body model represents only the shape).

\subsection{Challenge 2: Recovery of Generic Object Scans}

\begin{figure}
  \centering
  \includegraphics[width=\textwidth]{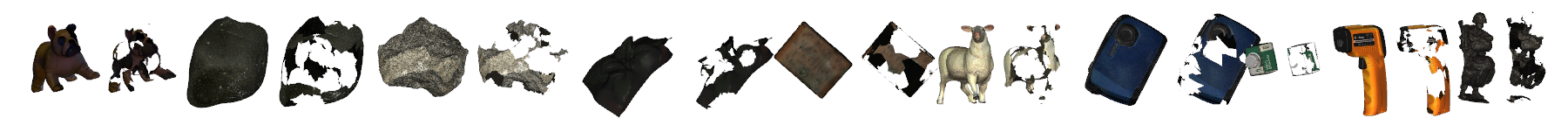}
  \includegraphics[width=\textwidth]{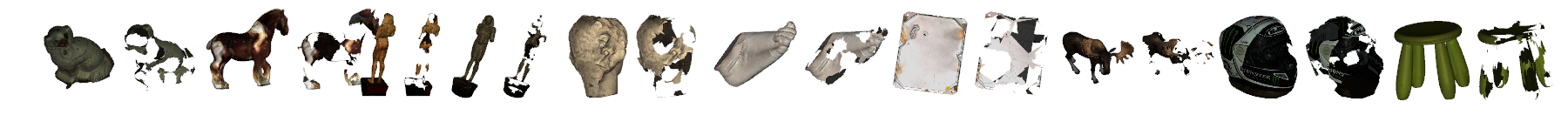}
  \includegraphics[width=\textwidth]{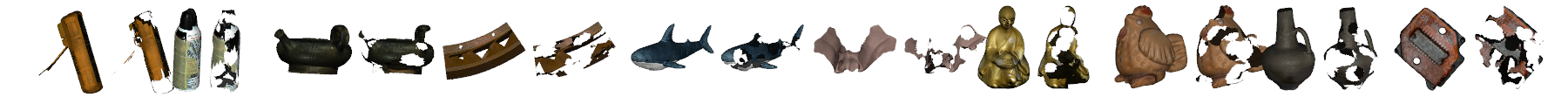}
  \caption{
    Samples of the 3DObjectTex dataset (normalised in scale for visualisation).
    The objects vary widely in scale and categories.
    For each object,
    ground-truth scan (left),
    sample synthetic partial scan (right).
  }
  \label{fig:3dobjecttex}
\end{figure}

Challenge 2 focuses on the shape and texture completion of generic objects.
Fig.~\ref{fig:3dobjecttex} shows some example objects.

\subsection{Partial data}

In all challenges, the partial data is generated synthetically by removing
surface regions randomly,
as shown in Fig.~\ref{fig:3dbodytex}~and Fig.~\ref{fig:3dobjecttex}.
This is done with a hole cutting operation following the steps below:\\
(1) take as input a central vertex $v_c$ and the number of vertices to remove $k$;\\
(2) select the $k$ vertices closest to $v_c$ in Euclidean distance;\\
(3) remove the selected vertices and adjacent triangles from the mesh.\\
The process is repeated 40 times, with $k$ set to 2\% of the points of the
mesh.
Most meshes in the datasets being regularly sampled, this is equivalent to
defining a nearest-neighbour radius proportional to the size of the mesh.
For challenge 1, the partial data is generated only in the considered regions
(everywhere except hands and head for Track 1, and only on the hands, ear and
nose in Track 2).
The partial data is generated by the participants themselves.
Additional methods for partial data generation were allowed if reported.
Routines to generate the partial data were distributed in the provided
software library.

\subsection{Evaluation data}

For evaluation, the partial data was generated by the organisers and shared with
the participants.
The reference data was held secret.

%% file: evaluation.tex
\section{Evaluation Metrics and Scores}
\label{sec:evaluation}

Entries to the challenge were ranked with an overall score, $S$,
that considers jointly
the quality of the shape and texture reconstructions,
and the amount of completeness.
It is based on the metrics proposed by Jensen~et~al.~\cite{jensen2014large},
with modifications and extensions for the task of shape and texture completion.
The main terms in computing the scores are
a surface-to-surface distance accounting for missing data in determining
correspondences,
a measure of the texture distance (on top of the shape distance),
surface area and hit-rate measures reflecting the degree of completion,
and a mapping function to convert distances to scores,
empirically adjusted from baseline data.
Below, the ground-truth mesh is denoted, $Y$,
and the estimated reconstruction, $X$.

\subsubsection{Overall score.}
A single reconstruction is scored with,
\begin{equation}
S = S_\alpha \frac{S_s + S_t}{2} \; \in [0, 1],
\label{equ:overall_score}
\end{equation}
where $S_a$, $S_s$ and $S_t$ are independent scores for
the surface area, shape and texture, respectively.
All scores take as input the reconstructed and the ground-truth meshes,
$S=S(X, Y)$,
and map to the interval $[0, 1]$.
A perfect reconstruction has a score of 1
and a poor reconstruction has a score tending towards 0.
A submission to the challenge contains reconstructions for all the samples
of the test set.
The submission is ranked by averaging the individual scores into a final global
score.

\subsubsection{Surface area score.}
The surface area score,
\begin{equation}
S_\alpha = 1-|\bar{A}_X - \bar{A}_Y| \; \in [0, 1],
\end{equation}
penalises a reconstruction proportionally to the deviation of its surface area
with respect to the ground truth.
$\bar{A}_X = \frac{A_X}{A_X + A_Y}$ is the normalised surface area of $X$
with respect to $Y$.
With a surface area lower or greater than the ground truth,
$S_a$ decreases proportionally the overall score~(\ref{equ:overall_score}).
As the surface area approaches the ground truth,
$S_a$ approaches 1, less affecting the overall score.
The surface area of a mesh is computed in practice by adding the surface areas
of all the individual triangles.

\subsubsection{Surface-to-surface distance.}
Both shape and texture scores, $S_s$ and $S_t$, are determined by estimating
the directed distances, $d^{XY}$ and $d^{YX}$,
between the ground truth mesh $Y$
and the estimated reconstruction $X$.

The directed distance between mesh $A$ and mesh $B$, $d^{AB}$, is computed by
sampling $N$ points uniformly on the surface of $A$,
finding the corresponding closest points on the surface of $B$,
and averaging the associated distances.

For a point $p^A\in\mathbb{R}^3$ on $A$, the corresponding closest point on $B$
is determined by computing the smallest point-to-triangle distance between
$p^A$ and all triangles of $B$.
The point-to-triangle distance, $d = d_0 + d_1$, is made of two components.
For a specific triangle in $B$,
$d_0$ is the Euclidean distance from $p^A$ to the closest point $p_0$
on the plane of the triangle.
$d_1$ is then the Euclidean distance from $p_0$ to the nearest point $p_1$
of the triangle (in the plane of the triangle).
If the intersection $p_0$ is inside the triangle, it is denoted as a
\emph{hit}, otherwise it is a \emph{miss}.
In case of a hit, $p_0 = p_1$, thus $d_1 = 0$ and $d = d_0$.

\subsubsection{Hit rate.}
When computing the surface-to-surface distance from mesh $A$ to mesh $B$,
the hit rate,
\begin{equation}
h^{AB} = \frac{H^{AB}}{N} \; \in [0, 1],
\end{equation}
is the proportion of the $N$ points sampled on $A$ \emph{hitting} $B$
(see previous paragraph).

\subsubsection{Shape score.}
The shape score,
\begin{equation}
  S_s =
    \frac{S_s^{XY} + S_s^{YX}}{2}
    =
    \frac{
      h^{XY} \phi_s(d_s^{XY})
      + h^{YX} \phi_s(d_s^{YX})
    }{2}
\end{equation}
is a measure of the similarity of the shape of two meshes.
The measure is symmetric by averaging the directed measures.
The hit rates, $h^{XY}$ and $h^{YX}$, penalise overcomplete and incomplete
reconstructions, respectively.
In the directed shape score,
\begin{equation}
S_s^{XY}=h^{XY} \phi_s(d_s^{XY}),
\end{equation}
the mapping function
$\phi_s: [0, \infty] \mapsto [0, 1]$
converts the computed distance to a score in $[0, 1]$.
It is defined by a normal distribution function with zero mean,
\begin{equation}
\phi_s(d)
 =
 \frac{1}{\sigma_s\sqrt{2\pi}}
 e^{-\frac{1}{2} {\left(\frac{d}{\sigma_s}\right)}^2},
\end{equation}
where the standard deviation $\sigma_s$ is estimated from baselines, including
the ground-truth data,
the input partial data,
additional perturbations thereof with local and/or global white Gaussian noise
on the shape,
and a baseline of shape reconstruction based on a hole-filling algorithm.

\subsubsection{Texture score.}
The texture score,
\begin{equation}
  S_t =
    \frac{S_t^{XY} + S_t^{YX}}{2}
    =
    \frac{
      h^{XY} \phi_t(d_t^{XY})
      + h^{YX} \phi_t(d_t^{YX})
    }{2}
\end{equation}
is similar in principle to the shape score, $S_s$, except that the
distance $d_t$ is computed in texture space
for all point correspondences obtained by the surface-to-surface distance.
Additionally, the parameter $\sigma_t$ for the mapping function $\phi_t$
specific to the texture
is estimated from
the ground-truth data,
the input partial data,
and additional perturbations with local and/or global white Gaussian noise,
in both shape and texture.
Below, the texture score is also interchangeably denoted colour score.

%% file: results.tex
\section{Results and Analysis}
\label{sec:results}

This section presents and analyses the submissions to both challenges.
The challenge has attracted 32 participants with 9 validated registrations for Challenge~1 and 6 for Challenge~2. Table~\ref{tab:submissions} gives
the number of valid submissions, received and accepted,
and the number of submitted solutions.

\begin{table}
\centering
\begin{tabular}{@{}llccc@{}}
 \toprule
 challenge & track & registrations & \multicolumn{2}{c}{submissions/participants} \\
 \cmidrule(lr){4-5}
 & & & received & accepted \\
 \midrule
 1 & 1 & 9 & 4/3 & 3/2 \\
 1 & 2 & 9 & 2/2 & 2/2 \\
 2 & \,- & 6 & 4/1 & 4/1 \\
 \bottomrule
 \\
\end{tabular}
\caption{
  Figures on validated registrations and entries for the challenges of SHARP 2020.
}
\label{tab:submissions}
\end{table}

\noindent The accepted entries are \emph{%
  Implicit Feature Networks for Texture Completion of 3D Data%
}~\cite{chibane2020cvpr,chibane2020implicit},
from RVH (Real Virtual Humans group at Max Planck Institute for Informatics),
submitted in several variants to both Challenge~1 and Challenge~2,
and
\emph{%
  3DBooSTeR: 3D Body Shape and Texture Recovery%
}~\cite{saint20203dbooster},
from SnT (Interdisciplinary Centre for Security, Reliability and Trust at the
University of Luxembourg),
submitted to Challenge~1.
In the following, the entries are interchangeably abbreviated
RVH-IF and SnT-3DB, respectively.
Table~\ref{tab:results} shows the quantitative results of the submissions for
both challenges.
The scores are reported in percent in an equivalent way to the scores mapping
to $[0, 1]$ in Section~\ref{sec:evaluation}.
The methods are compared to a baseline consisting of the unmodified partial
data.
The rest of this section presents and analyses the results.

\setlength{\tabcolsep}{6pt}
\begin{table}
  \centering
  \begin{tabular}{@{}llllll@{}}
    \toprule
    challenge & track & method & \multicolumn{3}{c}{score (\%)}\\
    \cmidrule(lr){4-6}
    & & & shape & texture & overall \\
    \midrule
    \multirow{4}{*}{1} & \multirow{4}{*}{1} & baseline & 38.95 $\pm$ 4.67 & 40.29 $\pm$ 4.24 & 39.62 $\pm$ 4.41 \\
    & & SnT-3DB & 54.21 $\pm$ 14.28 & 70.55 $\pm$ 7.26 & 62.38 $\pm$ 9.61\\
    & & RVH-IF-1 & 85.24 $\pm$ 5.72 & 87.69 $\pm$ 5.96 & 86.47 $\pm$ 5.38\\
    & & \textbf{RVH-IF-2} & \textbf{85.24 $\pm$ 5.72} & \textbf{88.26 $\pm$ 5.46} & \textbf{86.75 $\pm$ 5.19}\\
    \midrule
    \multirow{3}{*}{1} & \multirow{3}{*}{2} & baseline & 41.1 $\pm$ 3.31 & - & - \\
    & & SnT-3DB & 60.7 $\pm$ 10.98 & - & - \\
    & & \textbf{RVH-IF} & \textbf{83.0 $\pm$ 4.87} & - & - \\
    \midrule
    \multirow{5}{*}{2} & \multirow{5}{*}{\,-} & baseline & 42.09 $\pm$ 4.99 & 41.64 $\pm$ 5.24 & 41.87 $\pm$ 4.74 \\
    & & RVH-IF-1 & 72.68 $\pm$ 23.47 & 76.44 $\pm$ 14.33 & 74.56 $\pm$ 16.89 \\
    & & \textbf{RVH-IF-2} & \textbf{73.91 $\pm$ 22.85} & \textbf{76.93 $\pm$ 14.31} & \textbf{75.42 $\pm$ 16.68} \\
    & & RVH-IF-3 & 72.68 $\pm$ 23.47 & 53.13 $\pm$ 18.21 & 62.91 $\pm$ 16.09 \\
    & & RVH-IF-4 & 73.91 $\pm$ 22.85 & 53.73 $\pm$ 18.33 & 63.82 $\pm$ 16.01 \\
    \bottomrule
    \\
  \end{tabular}
  \caption{
    Reconstruction scores (\%) of the baseline and the proposed methods for
    both challenges.
    The baseline is the unmodified partial data.
  }
  \label{tab:results}
\end{table}

\subsection{Challenge 1 - Track 1}
\label{sec:results_1_1}

\begin{figure}[h]
  \centering
  \begin{subfigure}[b]{0.32\textwidth}
    \includegraphics[width=\textwidth]{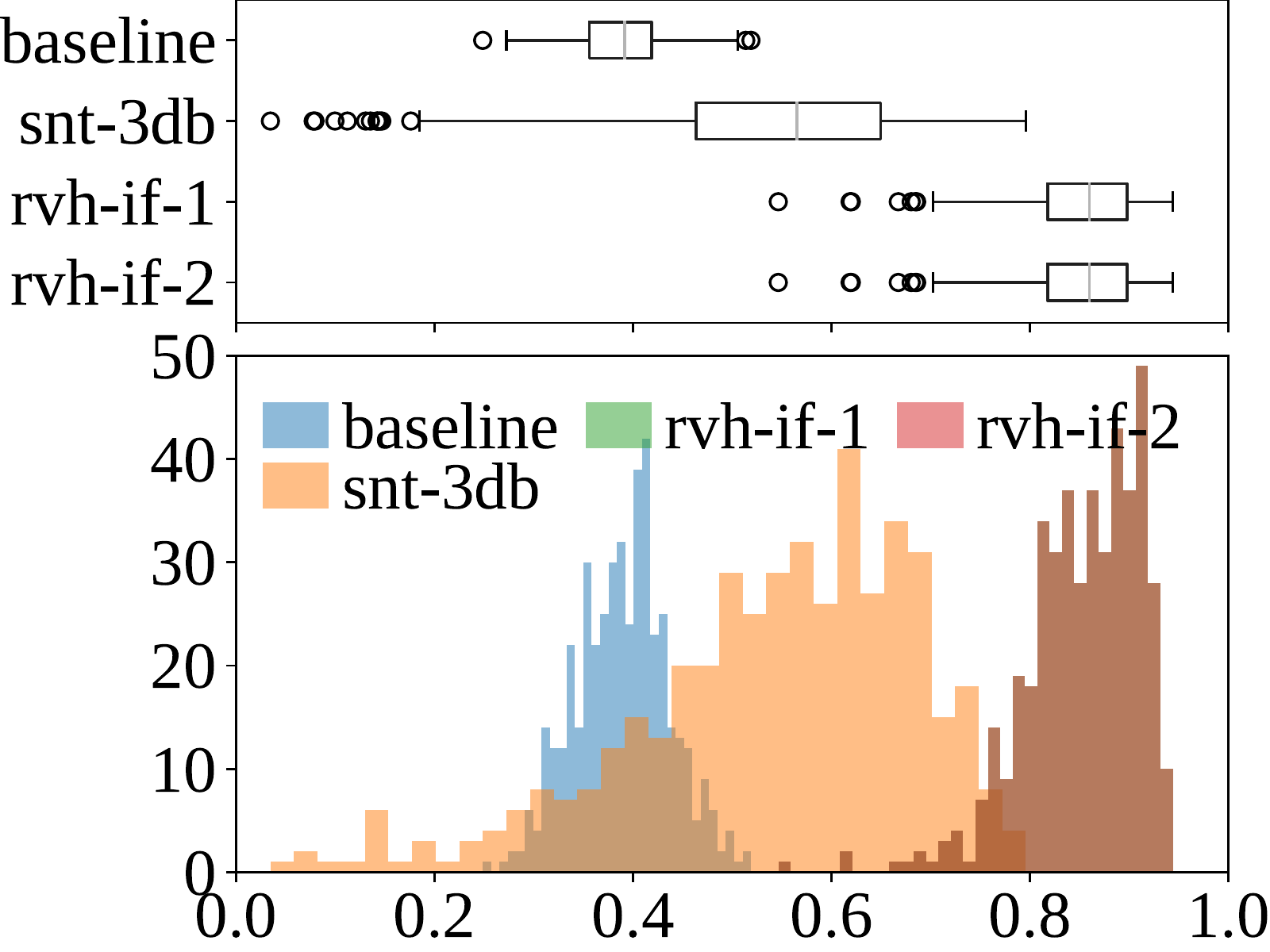}
    \caption{shape score}
  \end{subfigure}
  \begin{subfigure}[b]{0.32\textwidth}
    \includegraphics[width=\textwidth]{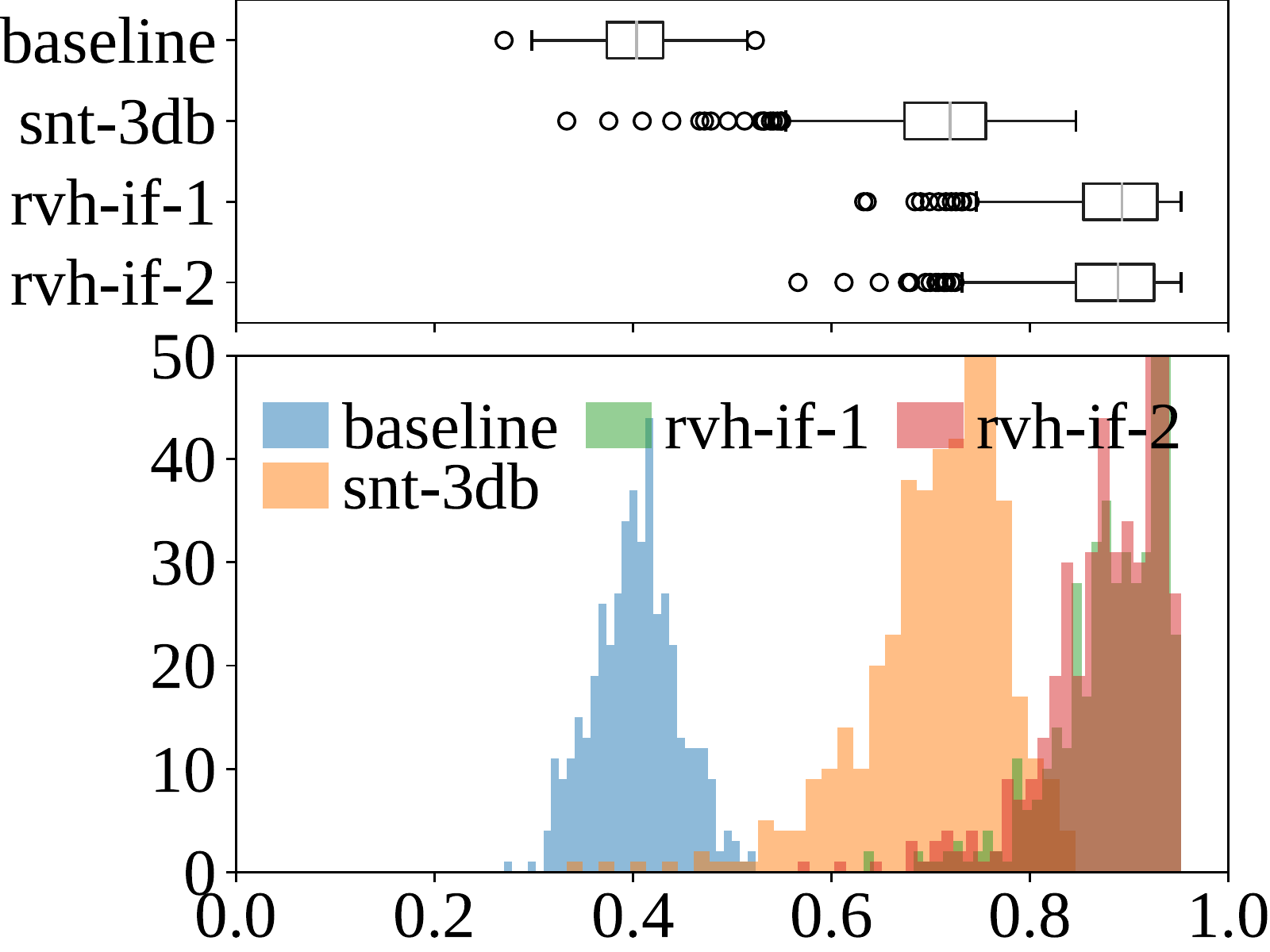}
    \caption{texture score}
  \end{subfigure}
  \begin{subfigure}[b]{0.32\textwidth}
    \includegraphics[width=\textwidth]{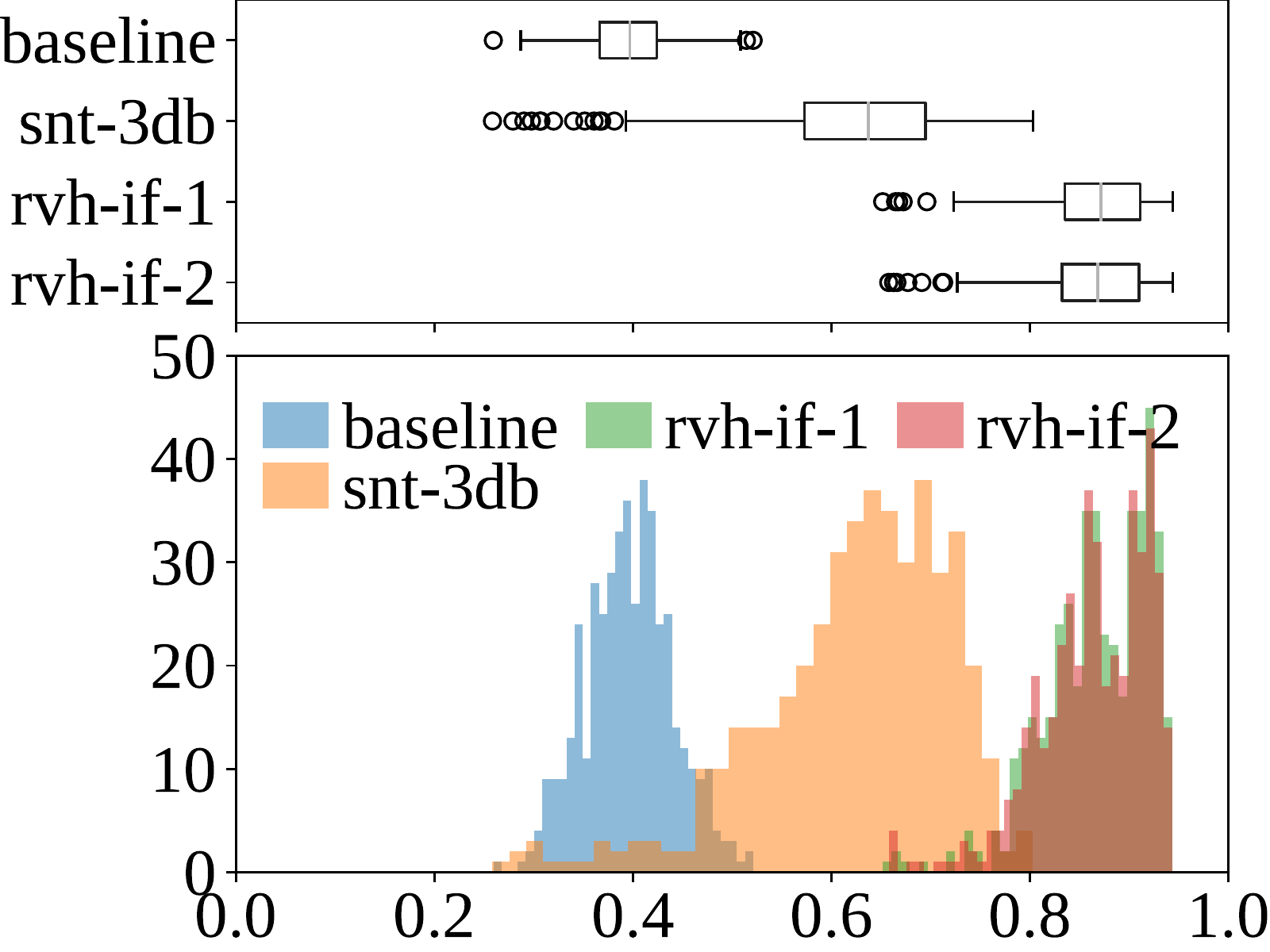}
    \caption{overall score}
  \end{subfigure}
\caption{
  Challenge 1 - track 1:
  Boxplots and frequency distributions of the reconstruction scores
  of the samples of the test set,
  for the baseline unmodified partial data,
  and all submissions, SnT-3DB and RVH-IF-\{1,2\}.
}
\label{fig:score_distribution_1_1}
\end{figure}

In challenge 1, track 1, the RVH-IF~\cite{chibane2020cvpr,chibane2020implicit}
approach ranks first with an overall reconstruction score of $86.75\%$
(see Table~\ref{tab:results}).
SnT-3DB comes second with $62\%$.
RVH-IF surpasses the baseline unmodified partial data with an overall score of
46\% higher
and performs significantly better than SnT-3DB~\cite{saint20203dbooster}
with a score increment of 24\%.
RVH-IF-2 has similar shape and texture scores with differences of 2-3\%,
while SnT-3DB has a much higher texture score, 16\% above the shape score.
The RVH-IF-2 variant slightly improves the texture score over RVH-IF-1,
with $88.26$\% instead of $87.69$\%, but the shape scores are identical.\\
Fig.~\ref{fig:score_distribution_1_1} shows the frequency distribution of the
scores.
RVH-IF appears relatively tightly located around the mean score 85\%
with a slight tail expanding in the lower scores down to 60\%.
SnT-3DB appears more spread for the shape score and with a tail reaching very low
scores down to 10\% below the baseline.

\begin{figure}[h]
  \centering
  \includegraphics[width=.32\textwidth]{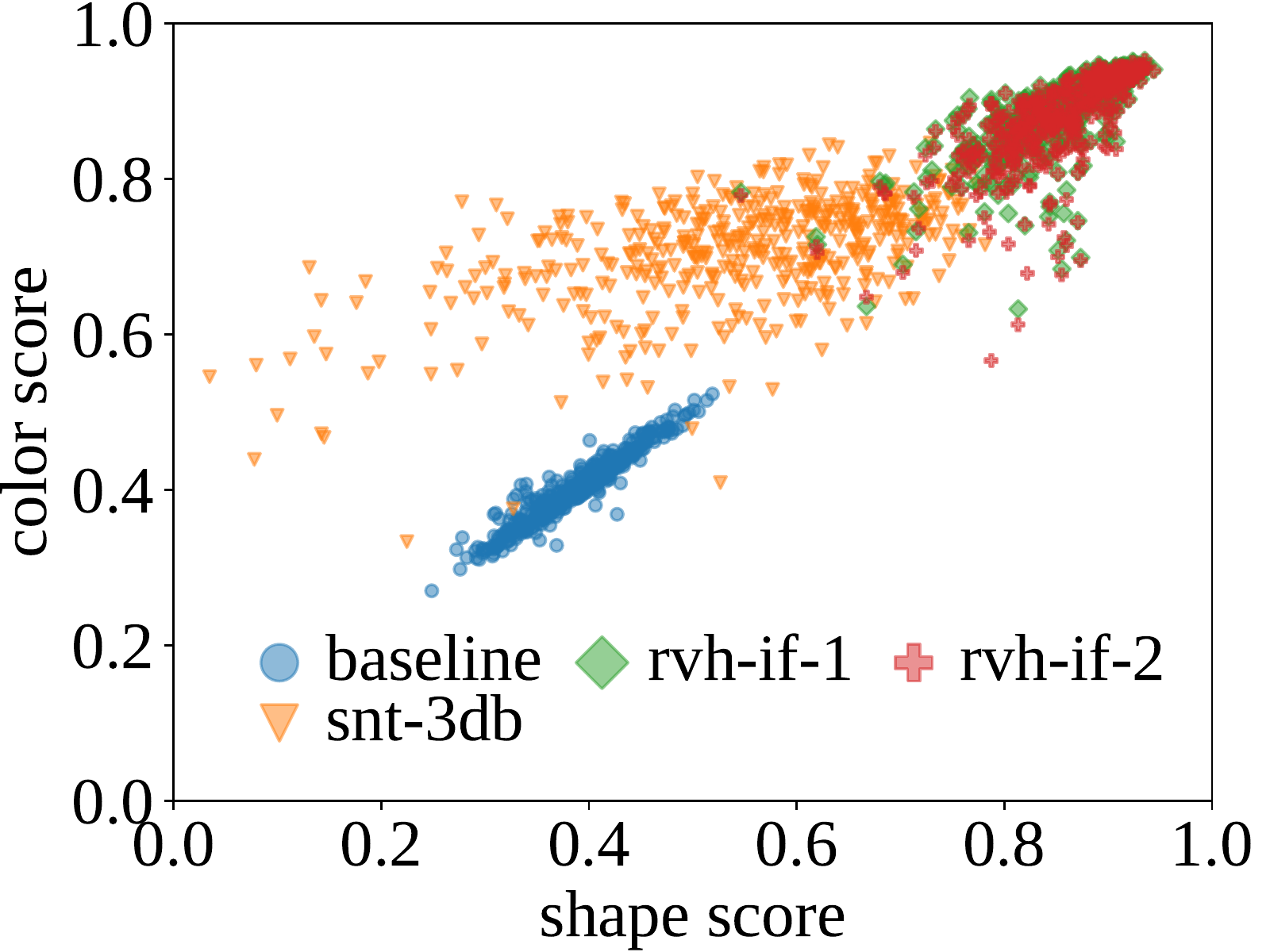}
  \caption{
    Challenge 1 - track 1:
    Correlation between shape and texture scores,
    for the baseline unmodified partial data,
    and all submissions, SnT-3DB and RVH-IF-\{1,2\}.
  }
  \label{fig:correlation_shape_texture}
\end{figure}

\begin{figure}
  \centering
  \begin{subfigure}[b]{0.24\textwidth}
    \includegraphics[width=\textwidth]{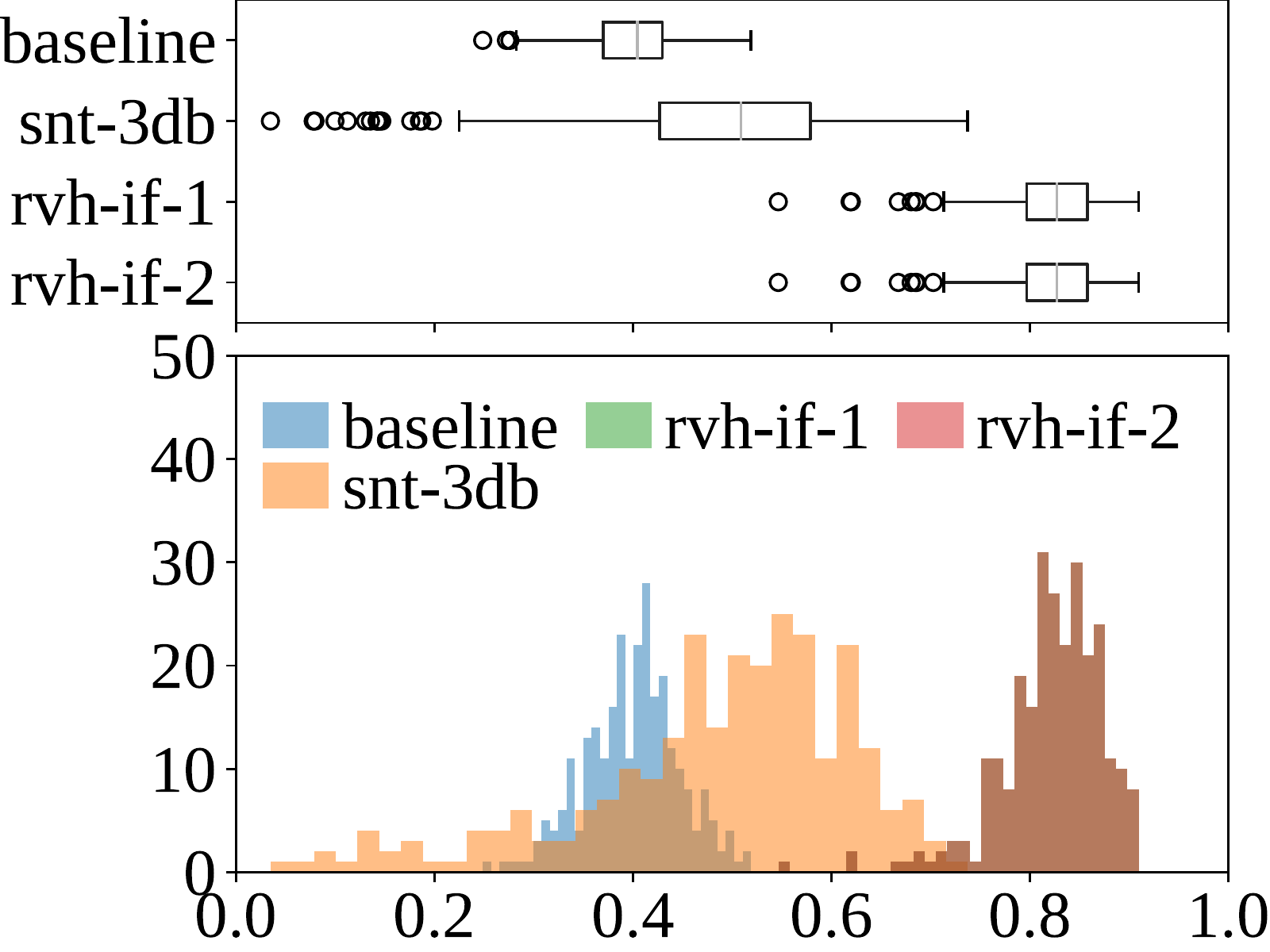}
  \end{subfigure}
  \begin{subfigure}[b]{0.24\textwidth}
    \includegraphics[width=\textwidth]{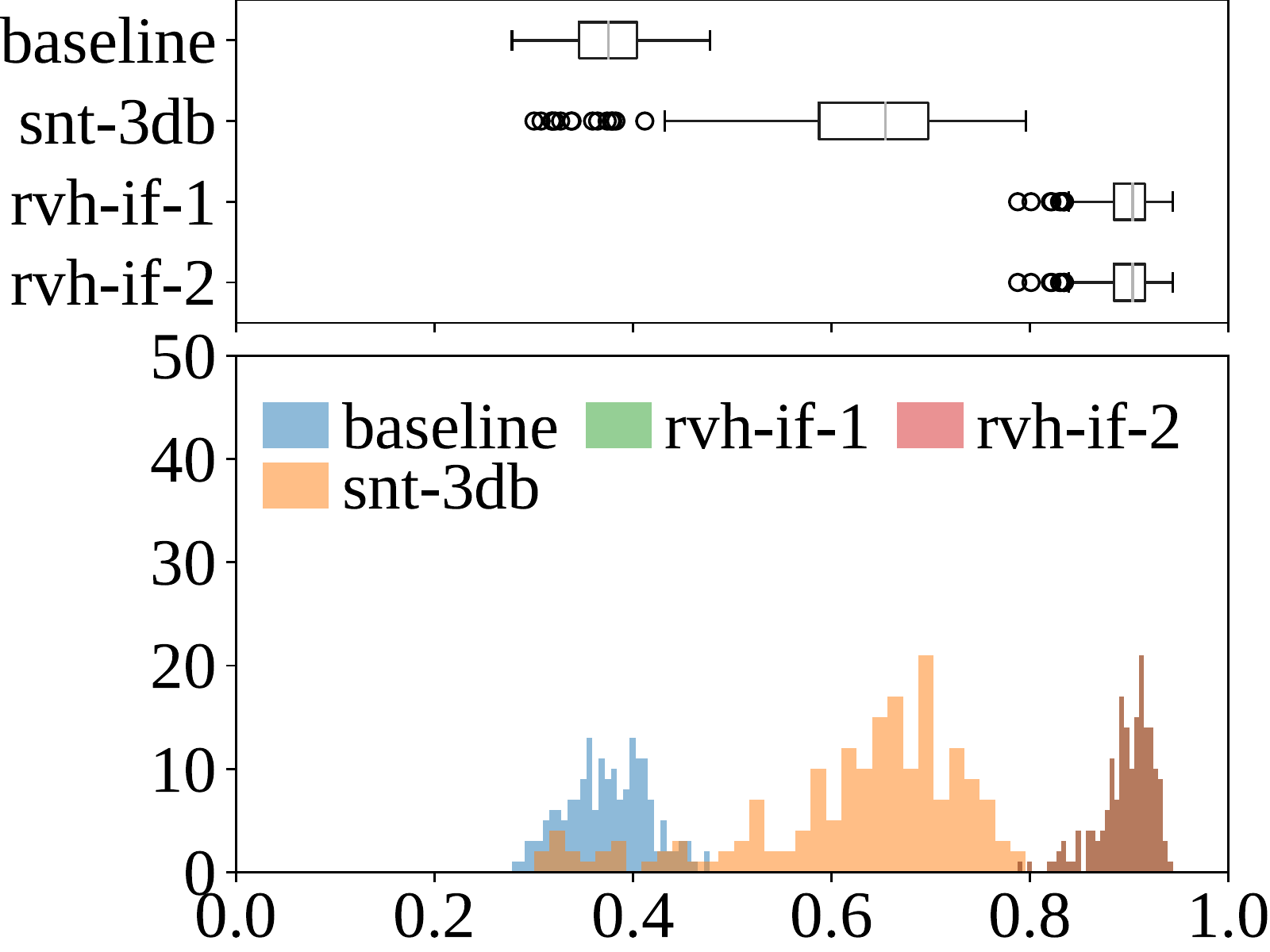}
  \end{subfigure}
  \begin{subfigure}[b]{0.24\textwidth}
    \includegraphics[width=\textwidth]{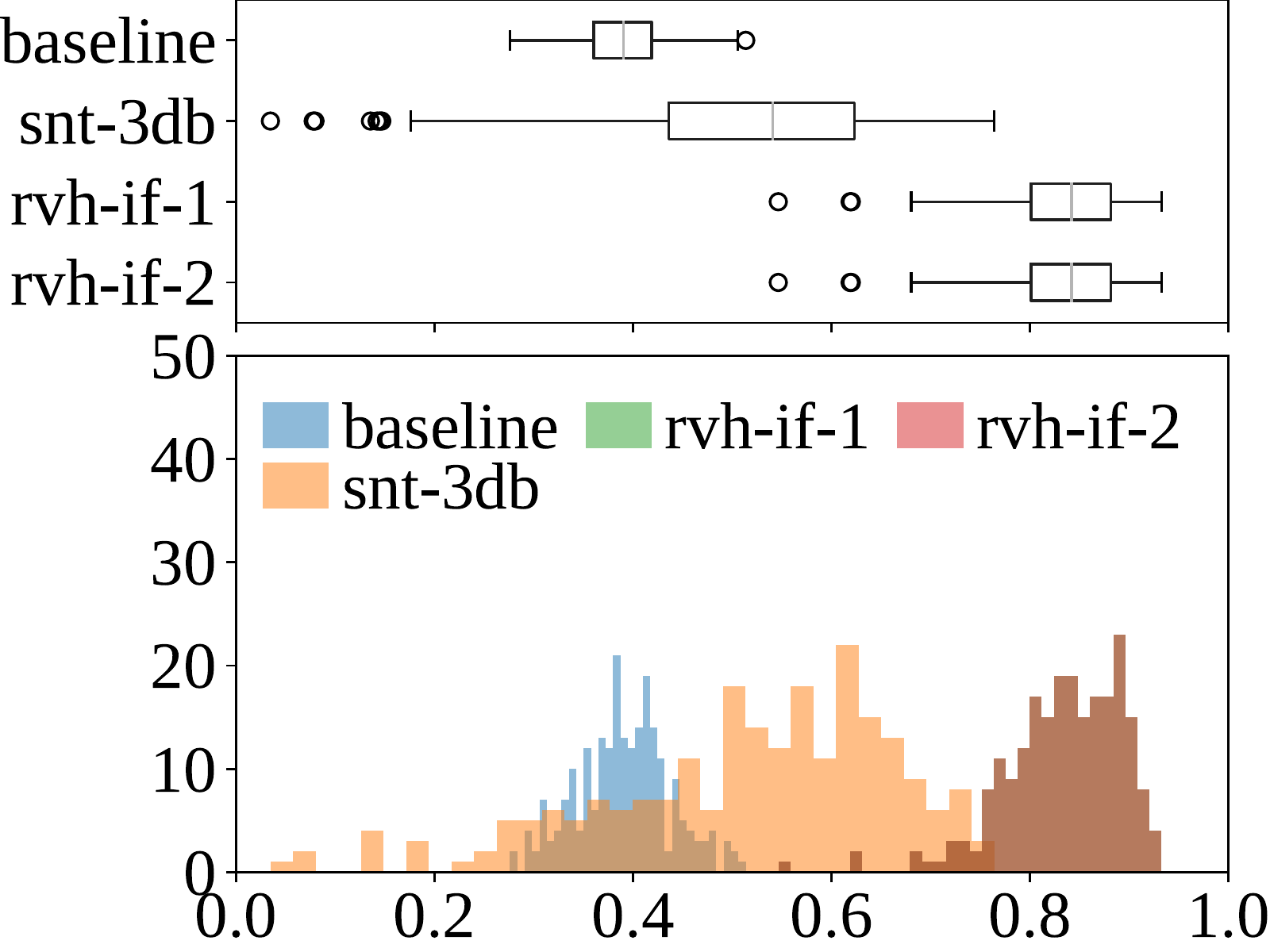}
  \end{subfigure}
  \begin{subfigure}[b]{0.24\textwidth}
    \includegraphics[width=\textwidth]{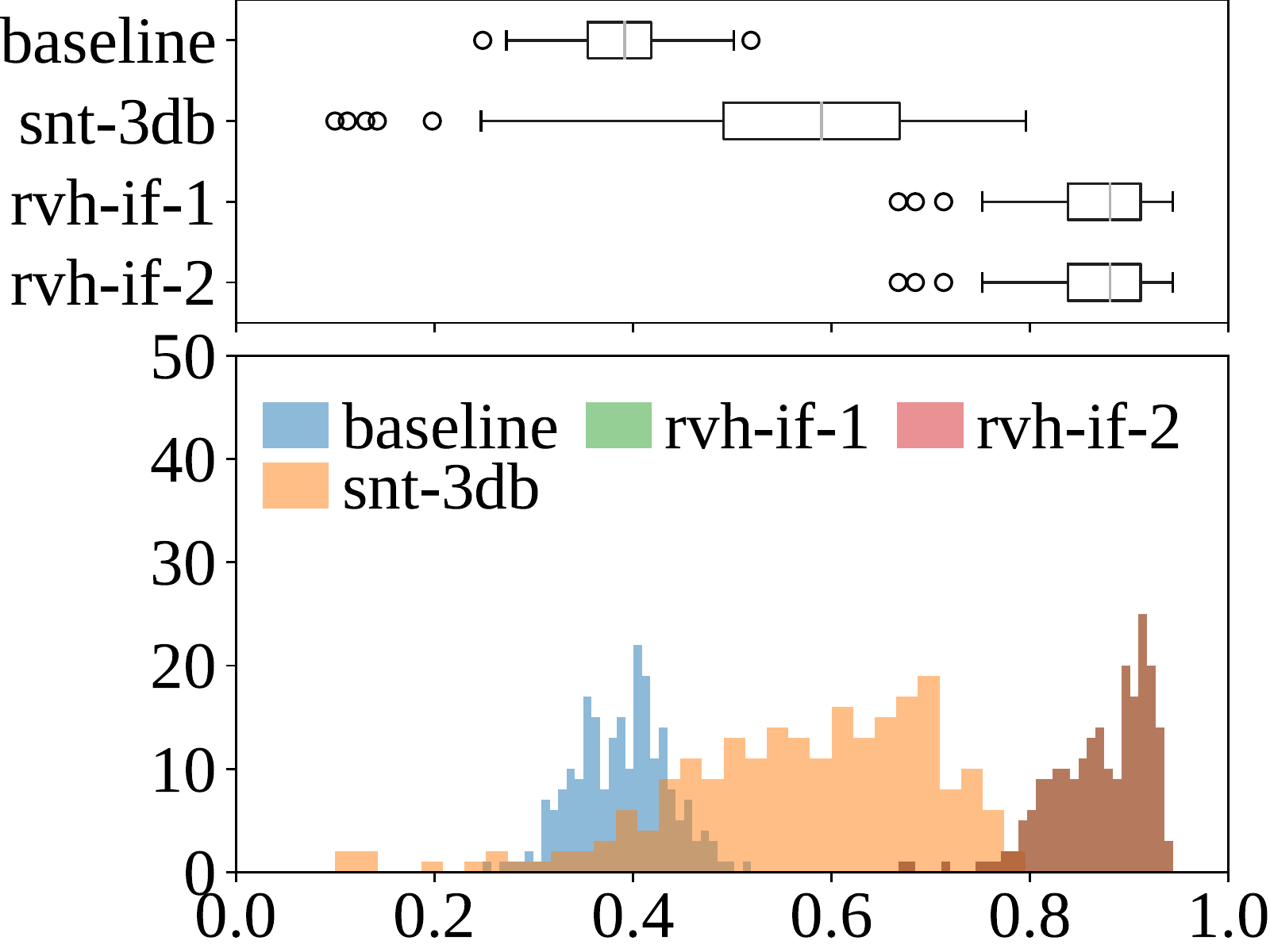}
  \end{subfigure}
  \\
  \begin{subfigure}[b]{0.24\textwidth}
    \includegraphics[width=\textwidth]{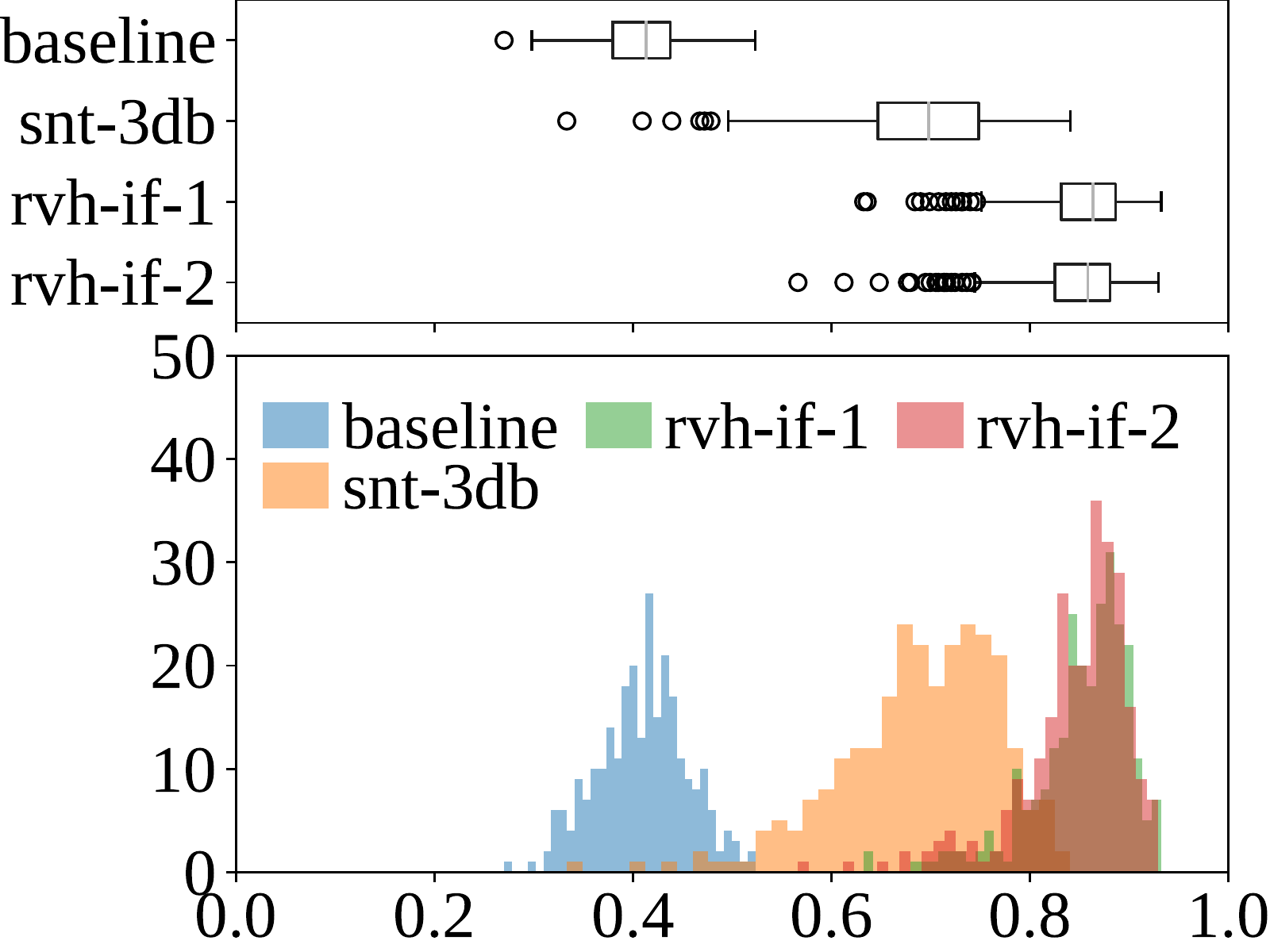}
  \end{subfigure}
  \begin{subfigure}[b]{0.24\textwidth}
    \includegraphics[width=\textwidth]{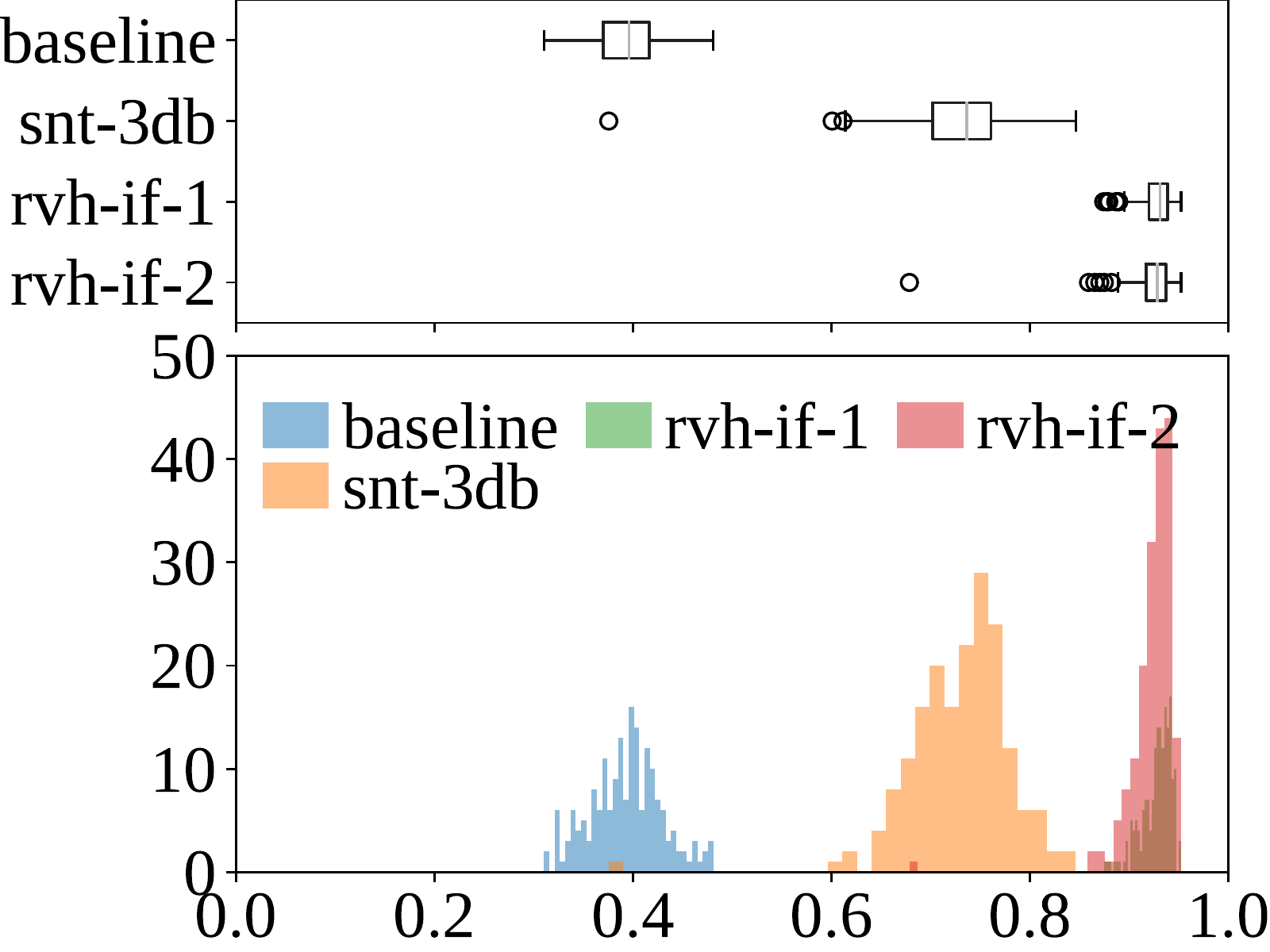}
  \end{subfigure}
  \begin{subfigure}[b]{0.24\textwidth}
    \includegraphics[width=\textwidth]{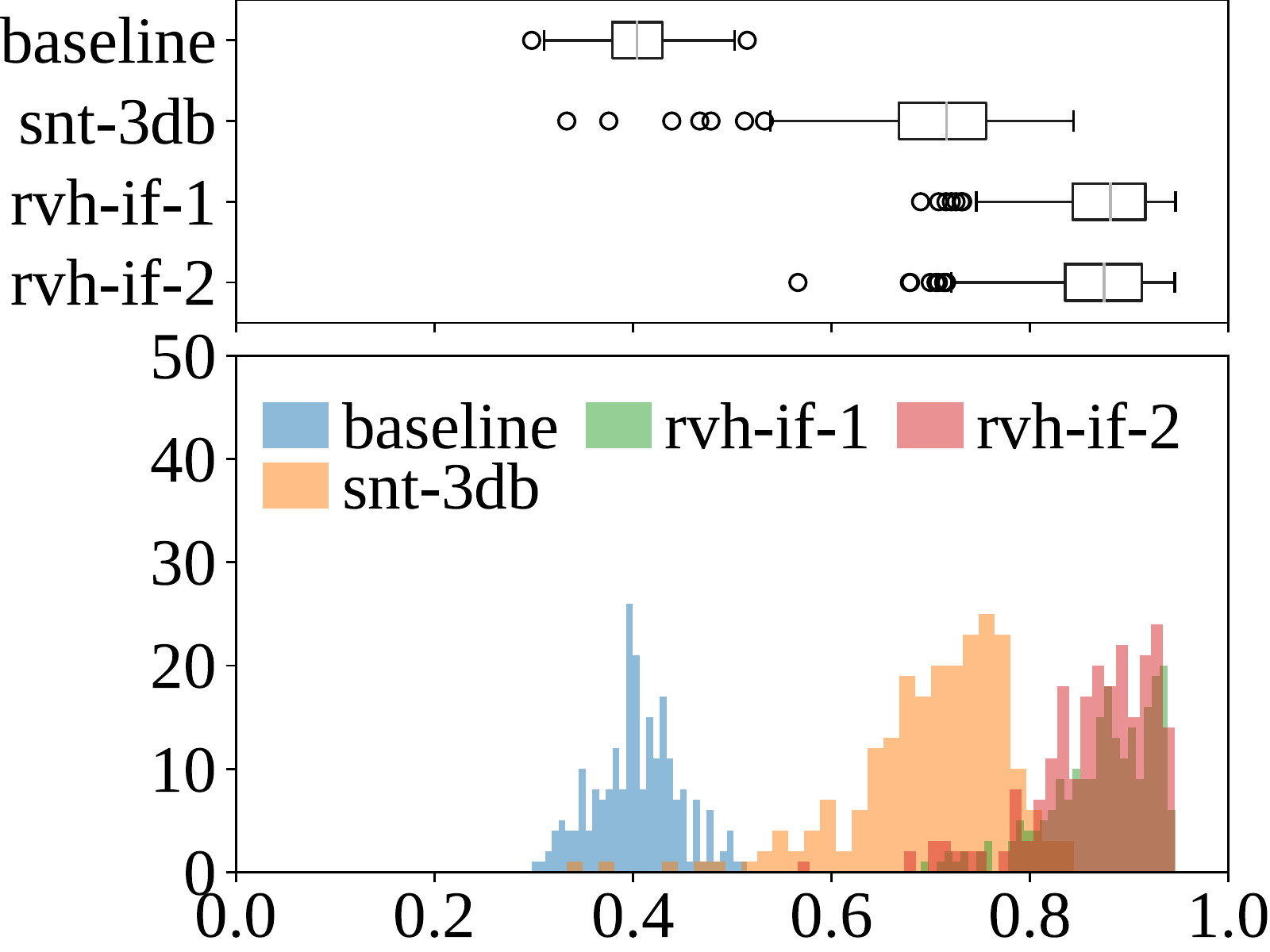}
  \end{subfigure}
  \begin{subfigure}[b]{0.24\textwidth}
    \includegraphics[width=\textwidth]{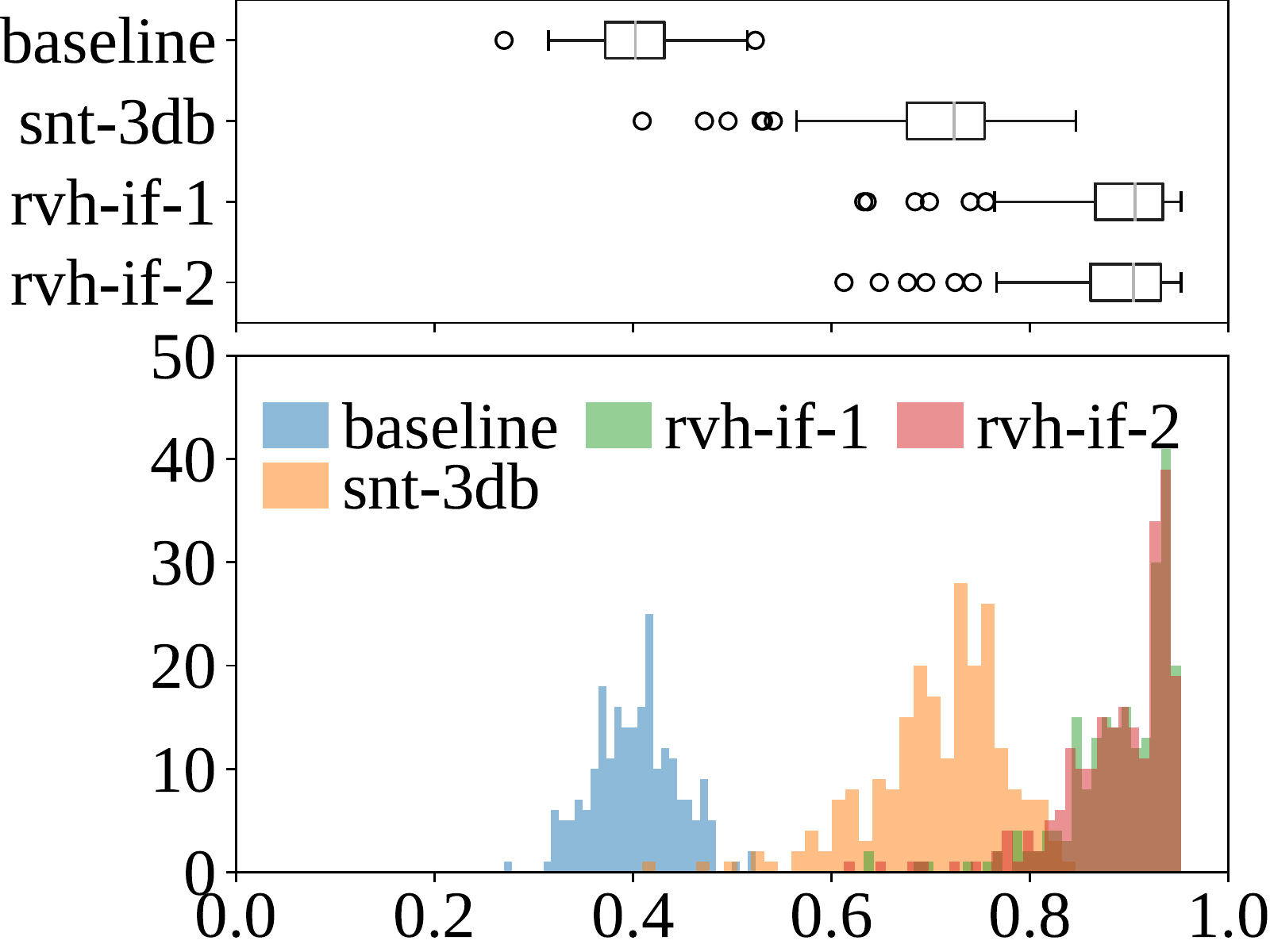}
  \end{subfigure}
  \\
  \begin{subfigure}[b]{0.24\textwidth}
    \includegraphics[width=\textwidth]{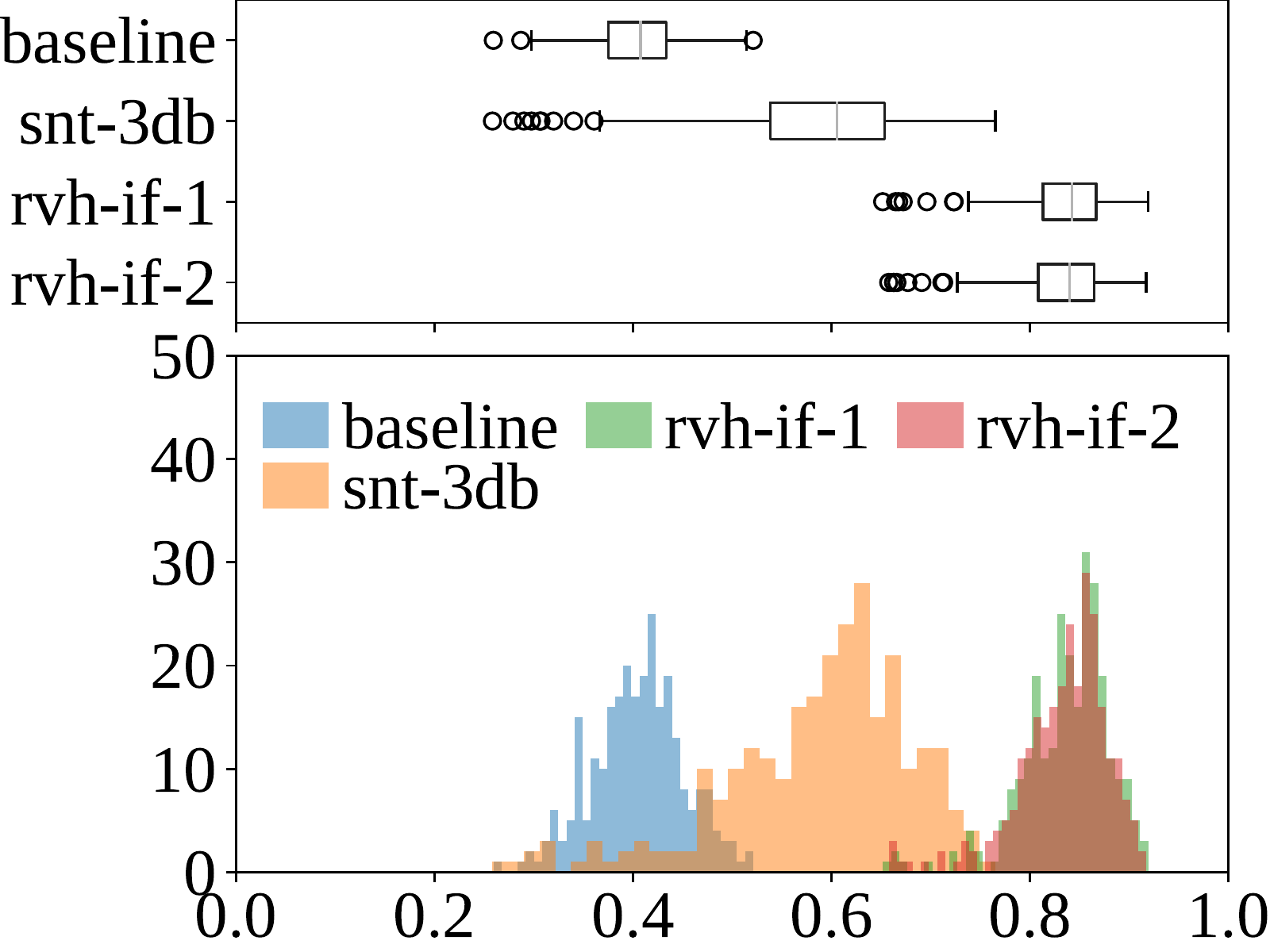}
  \end{subfigure}
  \begin{subfigure}[b]{0.24\textwidth}
    \includegraphics[width=\textwidth]{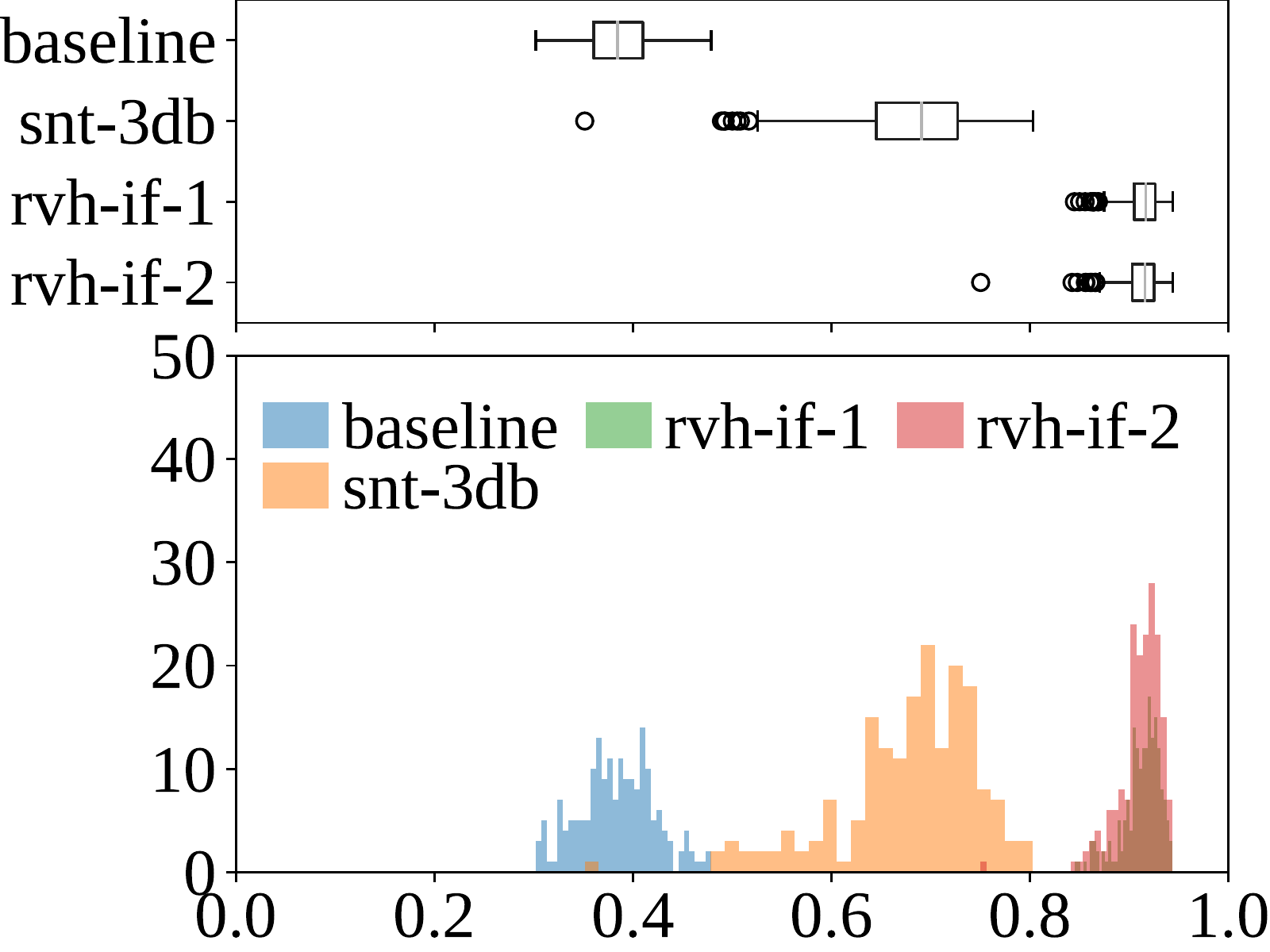}
  \end{subfigure}
  \begin{subfigure}[b]{0.24\textwidth}
    \includegraphics[width=\textwidth]{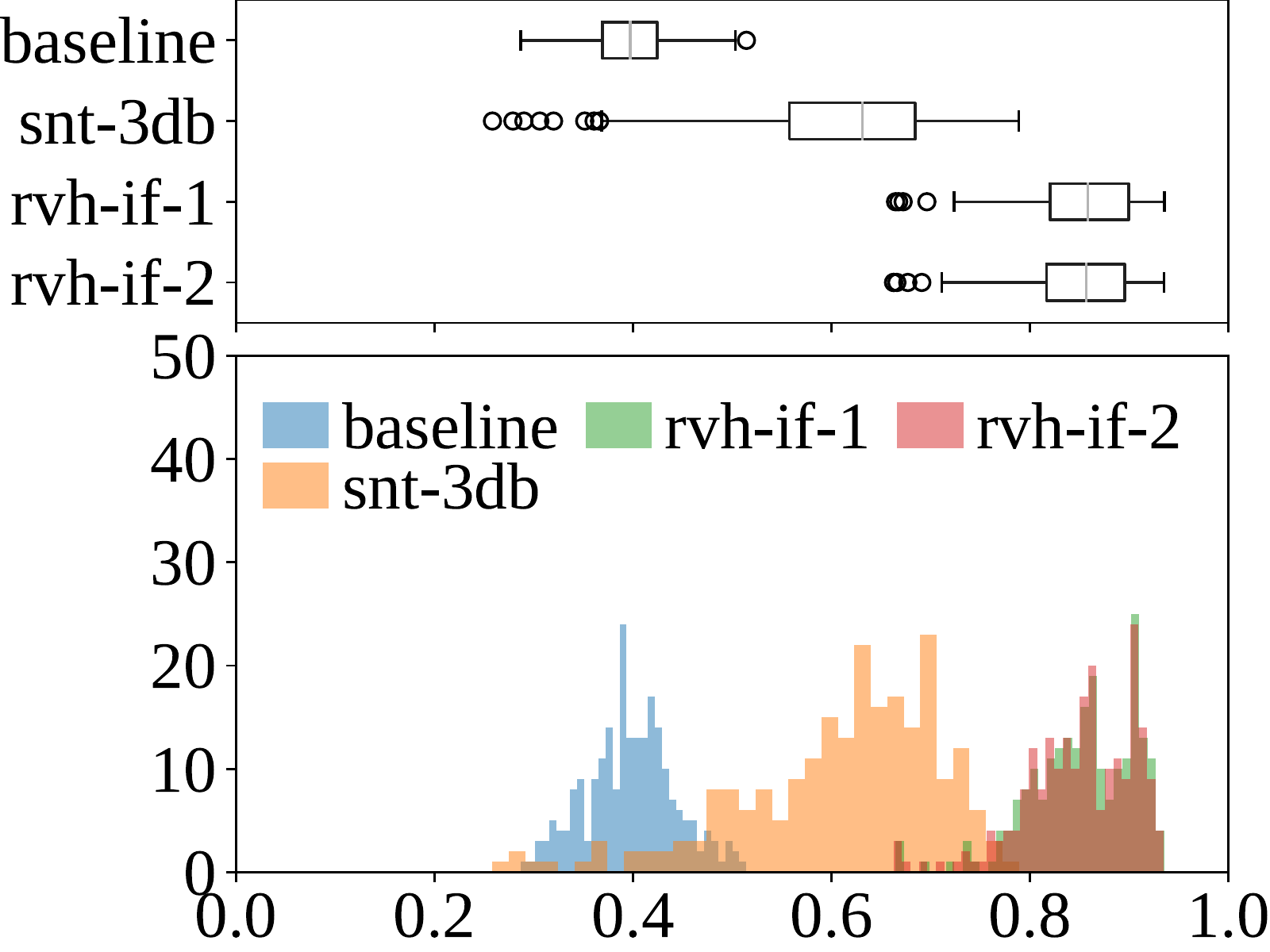}
  \end{subfigure}
  \begin{subfigure}[b]{0.24\textwidth}
    \includegraphics[width=\textwidth]{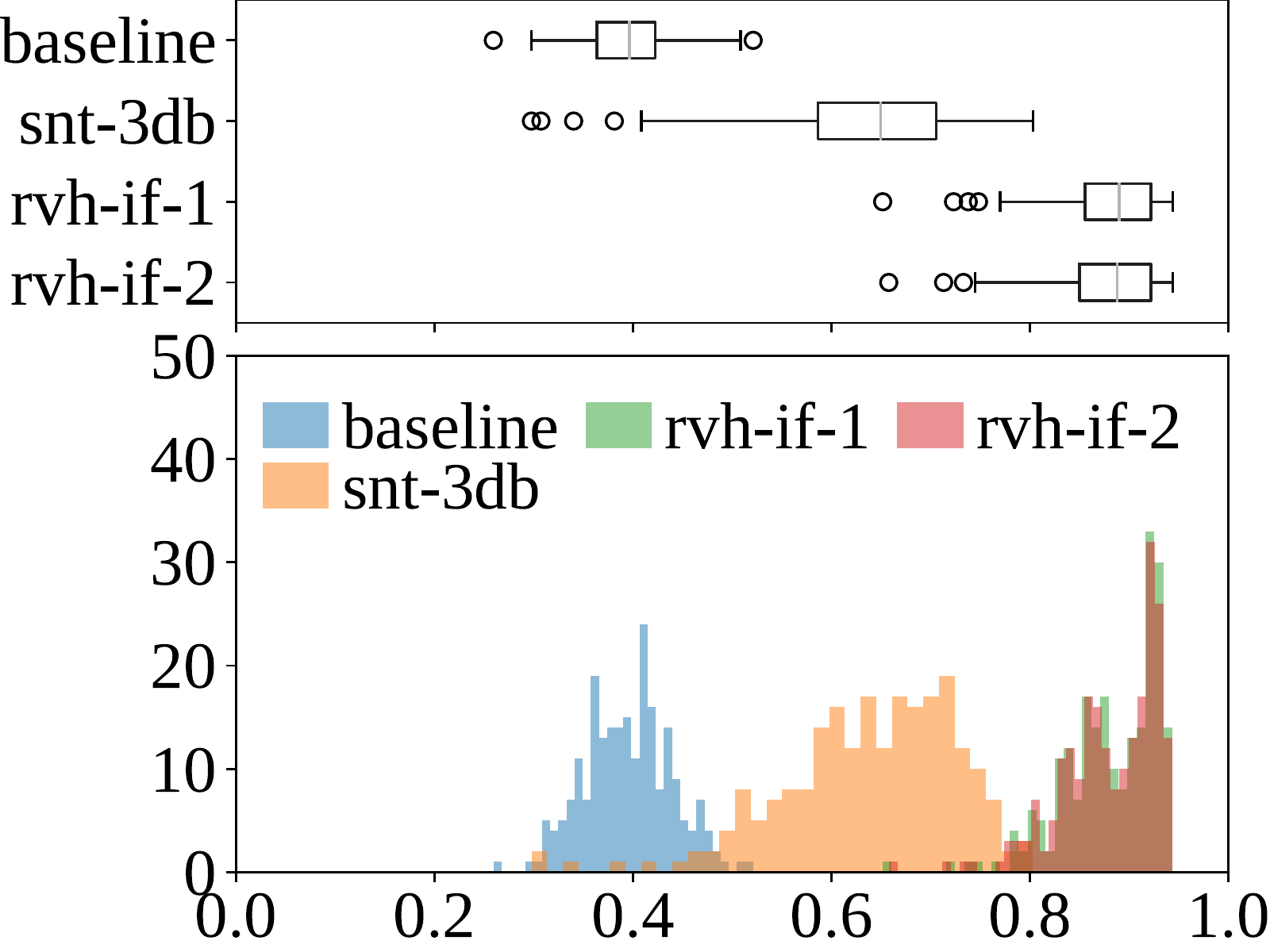}
  \end{subfigure}
  \\
  \begin{subfigure}[b]{0.24\textwidth}
    \caption{casual}
    \label{fig:distribution_subset_casual}
  \end{subfigure}
  \begin{subfigure}[b]{0.24\textwidth}
    \caption{fitness}
    \label{fig:distribution_subset_fitness}
  \end{subfigure}
  \begin{subfigure}[b]{0.24\textwidth}
    \caption{non-standard}
    \label{fig:distribution_subset_nonstandard_poses}
  \end{subfigure}
  \begin{subfigure}[b]{0.24\textwidth}
    \caption{standard}
    \label{fig:distribution_subset_standard_poses}
  \end{subfigure}
\caption{
  Challenge 1 - track 1:
  Boxplots and frequency distributions of the
  shape (top),
  texture (middle)
  and overall (bottom) scores
  on the test set,
  for subsets of 3DBodyTex.v2,
  and for the baseline unmodified partial data
  and all submissions, SnT-3DB and RVH-IF-\{1,2\}.
  Subsets (columns):
  casual and fitness clothing,
  non-standard and standard (A and U) poses.
}
\label{fig:distribution_subsets}
\end{figure}

\begin{figure}
  \centering
  \begin{subfigure}[b]{0.24\textwidth}
    \includegraphics[width=\textwidth]{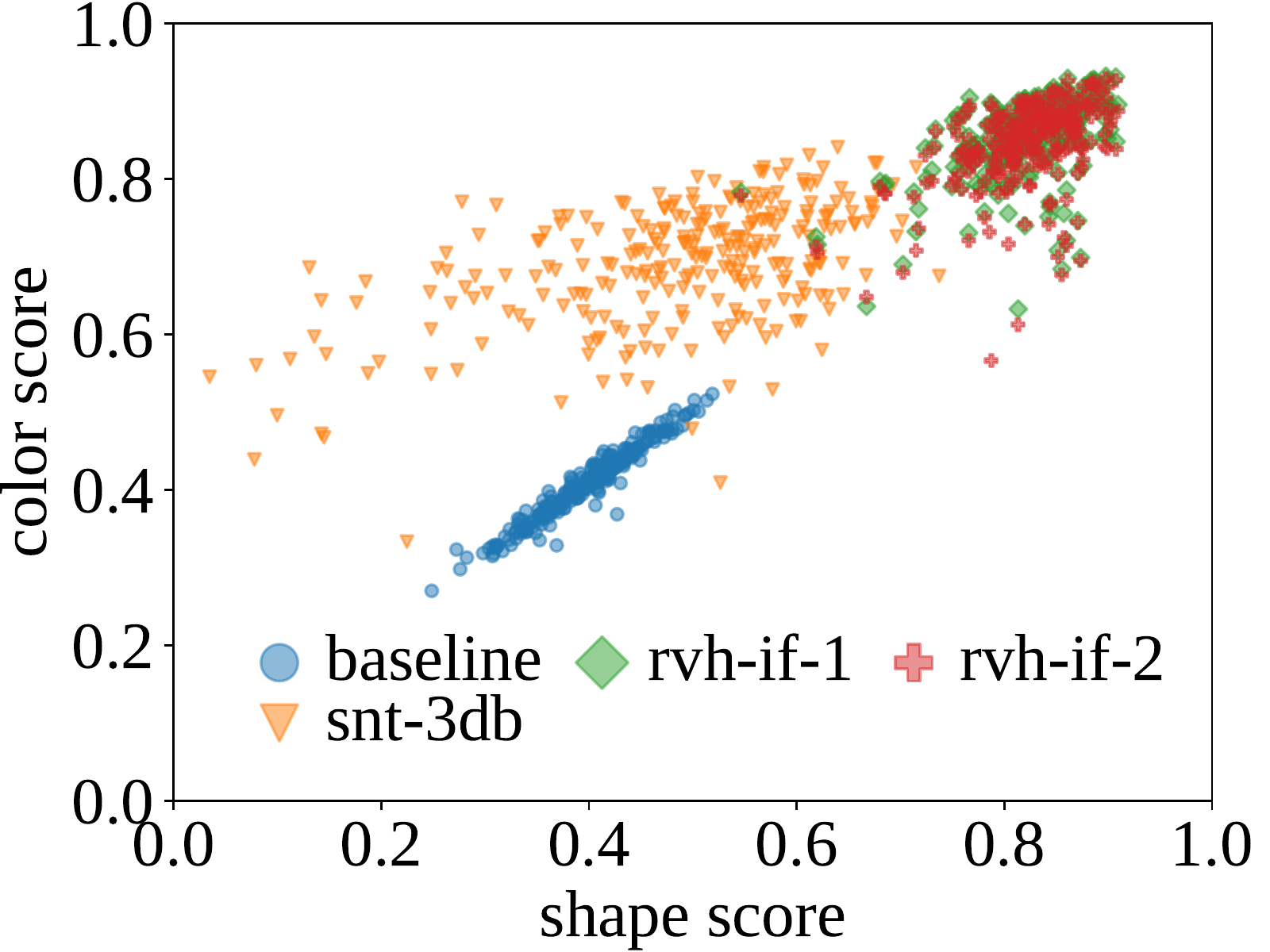}
  \end{subfigure}
  \begin{subfigure}[b]{0.24\textwidth}
    \includegraphics[width=\textwidth]{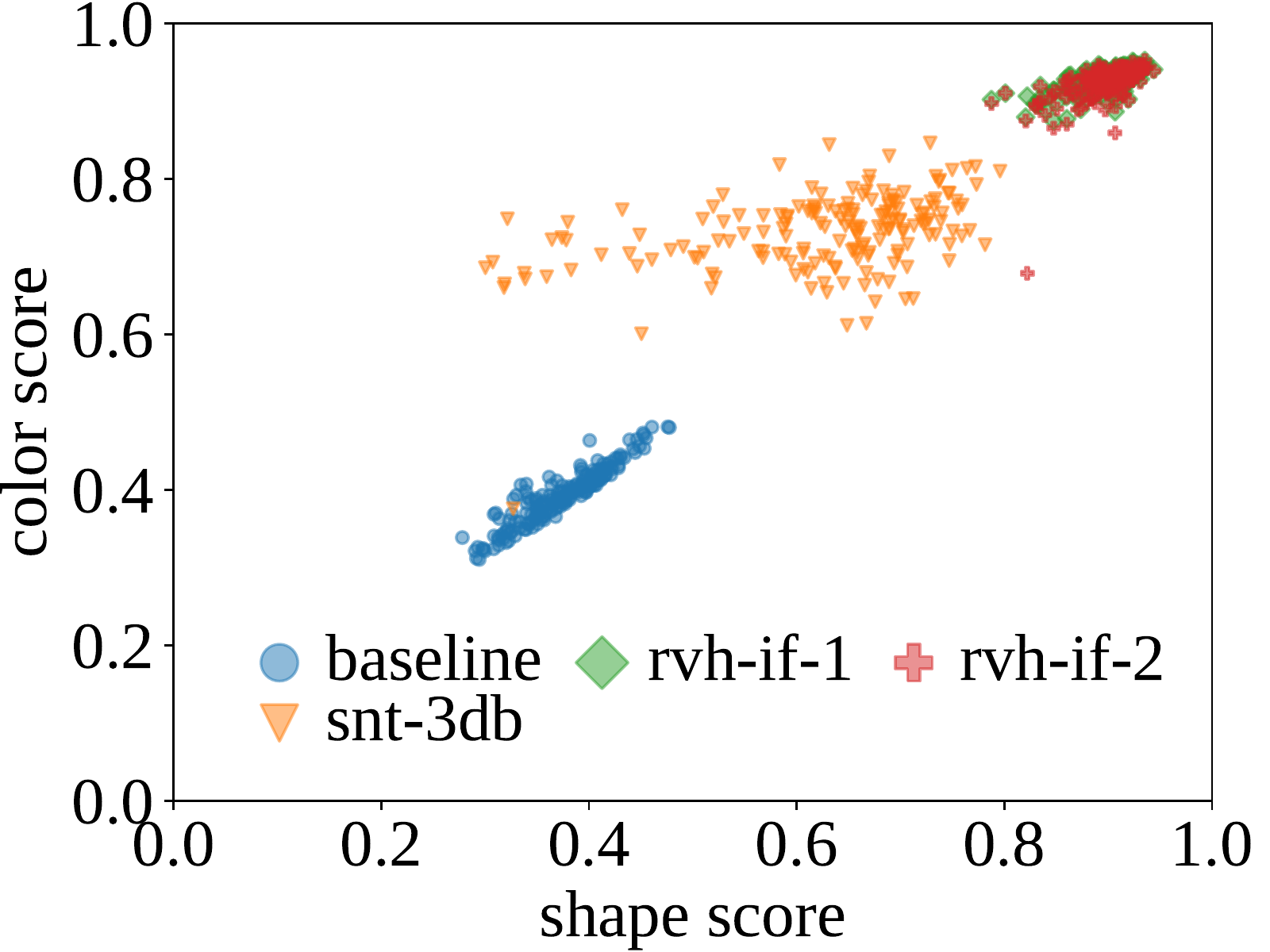}
  \end{subfigure}
  \begin{subfigure}[b]{0.24\textwidth}
    \includegraphics[width=\textwidth]{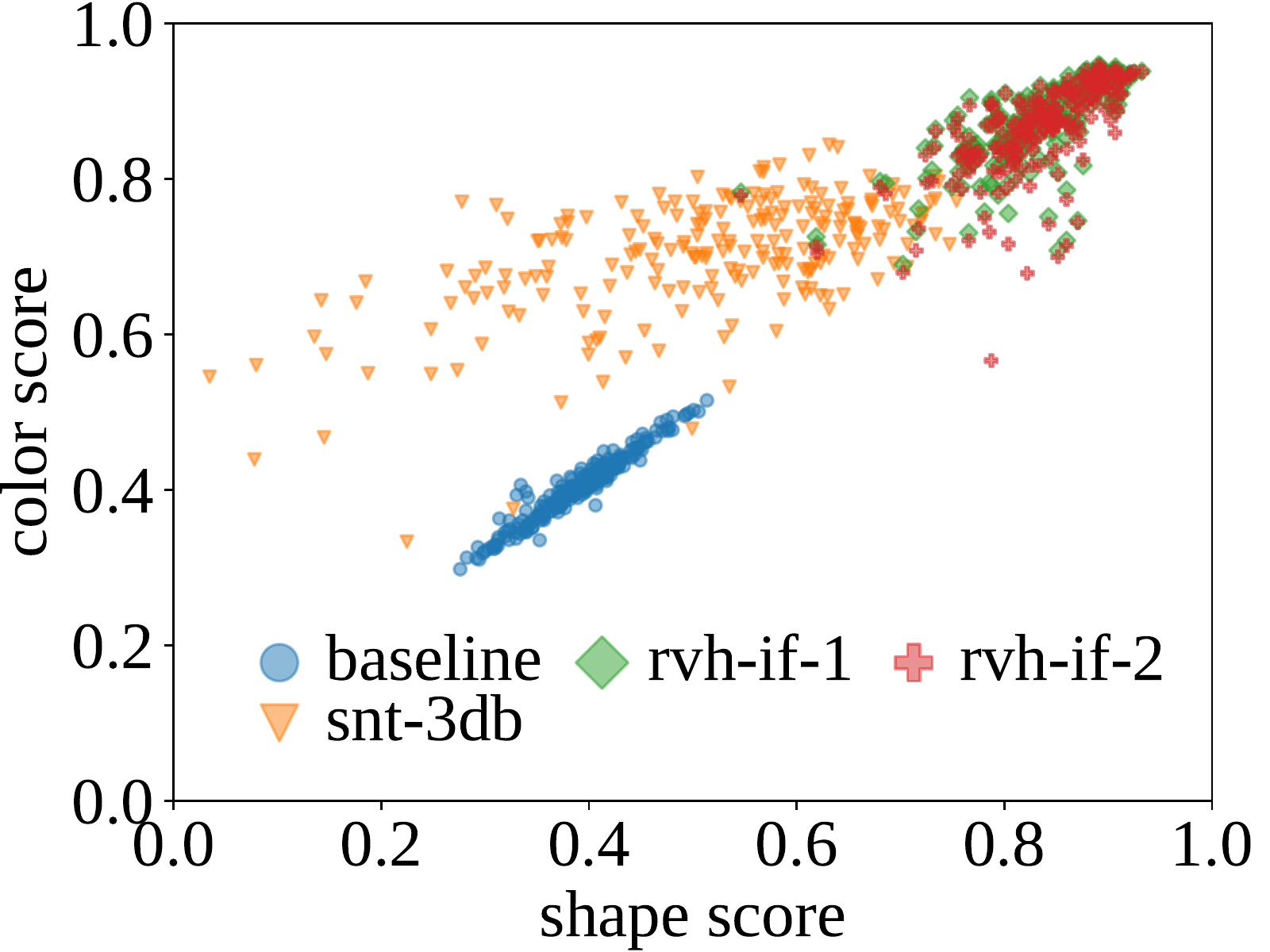}
  \end{subfigure}
  \begin{subfigure}[b]{0.24\textwidth}
    \includegraphics[width=\textwidth]{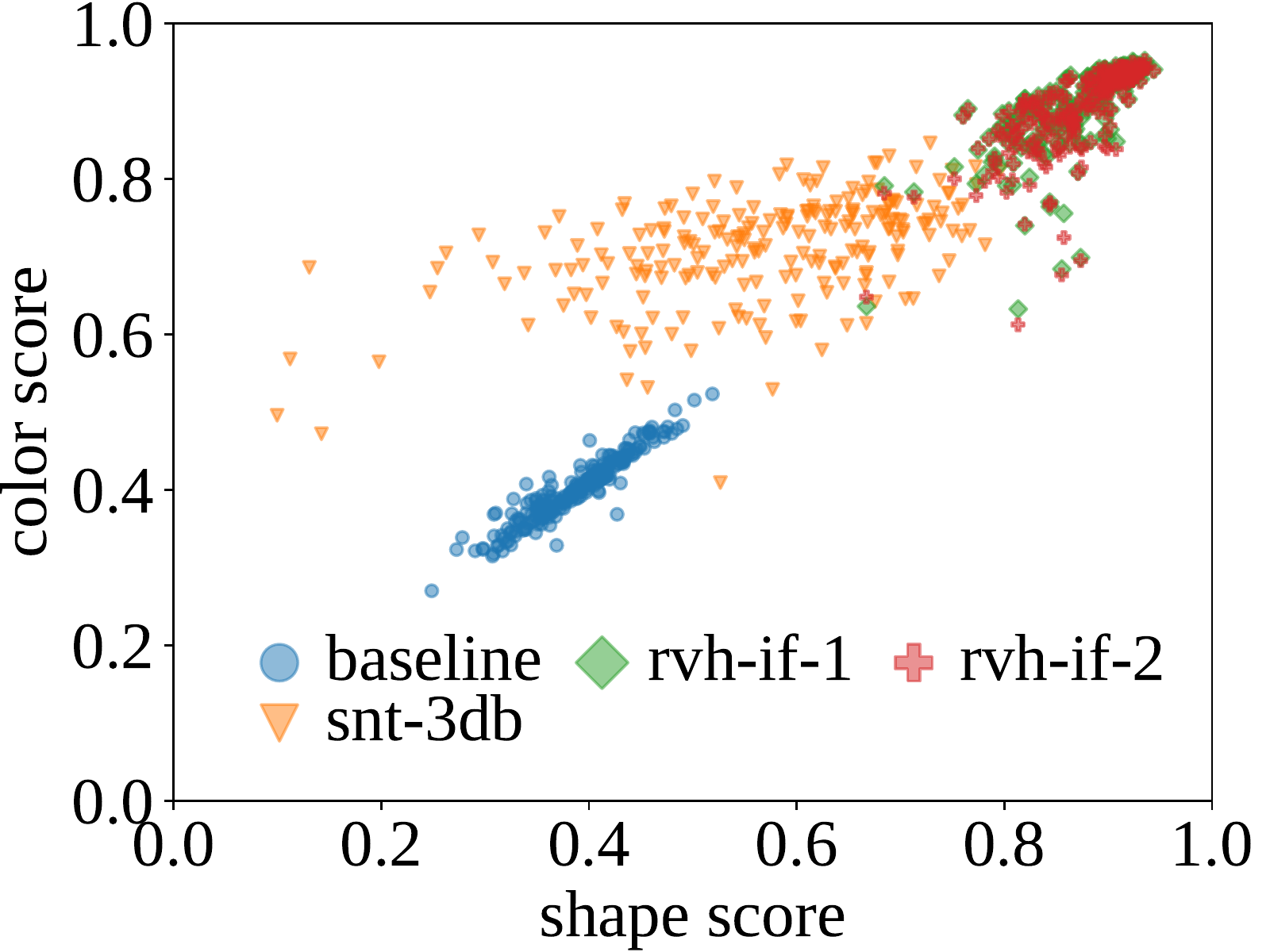}
  \end{subfigure}
  \\
  \begin{subfigure}[b]{0.24\textwidth}
    \caption{casual}
    \label{fig:correlation_subset_casual}
  \end{subfigure}
  \begin{subfigure}[b]{0.24\textwidth}
    \caption{fitness}
    \label{fig:correlation_subset_fitness}
  \end{subfigure}
  \begin{subfigure}[b]{0.24\textwidth}
    \caption{non-standard}
    \label{fig:correlation_subset_nonstandard_poses}
  \end{subfigure}
  \begin{subfigure}[b]{0.24\textwidth}
    \caption{standard}
    \label{fig:correlation_subset_standard_poses}
  \end{subfigure}
\caption{
  Challenge 1 - track 1:
  Correlation of shape and texture scores
  on the test set,
  for subsets of 3DBodyTex.v2,
  and for the baseline unmodified partial data
  and all submissions, SnT-3DB and RVH-IF-\{1,2\}.
  Subsets (columns):
  casual and fitness clothing,
  non-standard and standard (A and U) poses.
}
\label{fig:correlation_subsets}
\end{figure}

\noindent Fig.~\ref{fig:correlation_shape_texture} highlights the correlation of the
shape and texture scores.
The baseline unmodified partial data is highly correlated, which is expected
because the partial data available is exact.
RVH-IF methods display a high correlation, with overall a tendency towards
better texture scores than shape scores, except for a small proportion of
grouped outlier samples with higher shape and lower texture scores.
The SnT-3DB method displays the same tendency towards better texture than
shape but with more dispersed results.
Not a significant group of outliers is observed.
The weaker correlation in SnT-3DB could be due to at least two factors:
the sequential nature of the method, making the shape and texture reconstructions
somewhat independent of each other;
the mostly uniform colour of most clothing and of the body shape, resulting in
good texture recovery even if the spatial regions are not matched correctly.\\
Fig.~\ref{fig:distribution_subsets}~and~\ref{fig:correlation_subsets}
show the distribution of scores and the correlation between shape and texture
scores for subsets of the 3DBodyTex.v2 dataset.
The casual clothing subset (Fig.~\ref{fig:distribution_subset_casual})
shows lower and more dispersed scores than the fitness clothing
(Fig.~\ref{fig:distribution_subset_fitness}).
This reflects the diversity in texture and shape in the casual clothing,
which is challenging for all proposed methods.
Moreover, to a small but noticeable extent,
the reconstruction scores are higher
and more correlated in shape and texture on standard poses
(Fig.~\ref{fig:correlation_subset_standard_poses})
than on non-standard ones
(Fig.~\ref{fig:correlation_subset_nonstandard_poses}).

\begin{figure}
\centering
\begingroup 
\setlength{\tabcolsep}{3pt}
\renewcommand{\arraystretch}{1}
\newcommand\widthfactor{.09}
\begin{tabular}{@{}*{4}{c}@{}}
    \includegraphics[width=\widthfactor\textwidth]{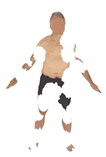}
  & \includegraphics[width=\widthfactor\textwidth]{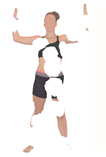}
  & \includegraphics[width=\widthfactor\textwidth]{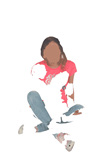}
  & \includegraphics[width=\widthfactor\textwidth]{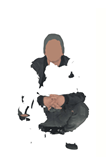}
  \\
    \includegraphics[width=\widthfactor\textwidth]{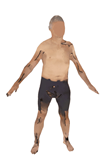}
  & \includegraphics[width=\widthfactor\textwidth]{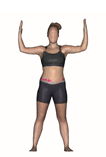}
  & \includegraphics[width=\widthfactor\textwidth]{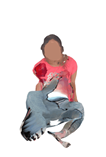}
  & \includegraphics[width=\widthfactor\textwidth]{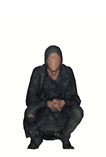}
  \\
    \includegraphics[width=\widthfactor\textwidth]{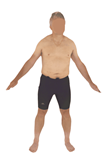}
  & \includegraphics[width=\widthfactor\textwidth]{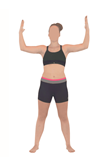}
  & \includegraphics[width=\widthfactor\textwidth]{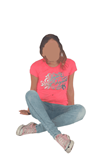}
  & \includegraphics[width=\widthfactor\textwidth]{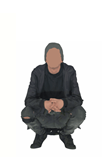}
  \\
  SnT-3DB & RVH-IF-1
  &
  SnT-3DB & RVH-IF-1
  \\
  \cmidrule(lr){1-2}
  \cmidrule(lr){3-4}
  \multicolumn{2}{c}{best}
  &
  \multicolumn{2}{c}{worst}
  \\
\end{tabular}
\endgroup
\caption{
  Challenge 1 - track 1:
  Best and worst reconstructions for submissions SnT-3DB and RVH-IF-1.
  From top to bottom:
  partial scan,
  reconstruction,
  ground truth.
}
\label{fig:best_worst_1_1}
\end{figure}

As a qualitative assessment, the best and worst reconstructions are shown
in Fig.~\ref{fig:best_worst_1_1} for both approaches SnT-3DB and the RVH-IF.
The best reconstructions are for scans in fitness clothing and standard pose,
as also observed quantitatively above.
The worst reconstructions are for scans in casual clothing and non-standard
sitting/crouching poses, containing more shape and texture variation.
It could also be that the ground truth for this data is less reliable, biasing
the results towards lower scores.

\subsection{Challenge 1 - Track 2}
\label{sec:results_1_2}

\begin{figure}
  \centering
  \includegraphics[width=.32\textwidth]{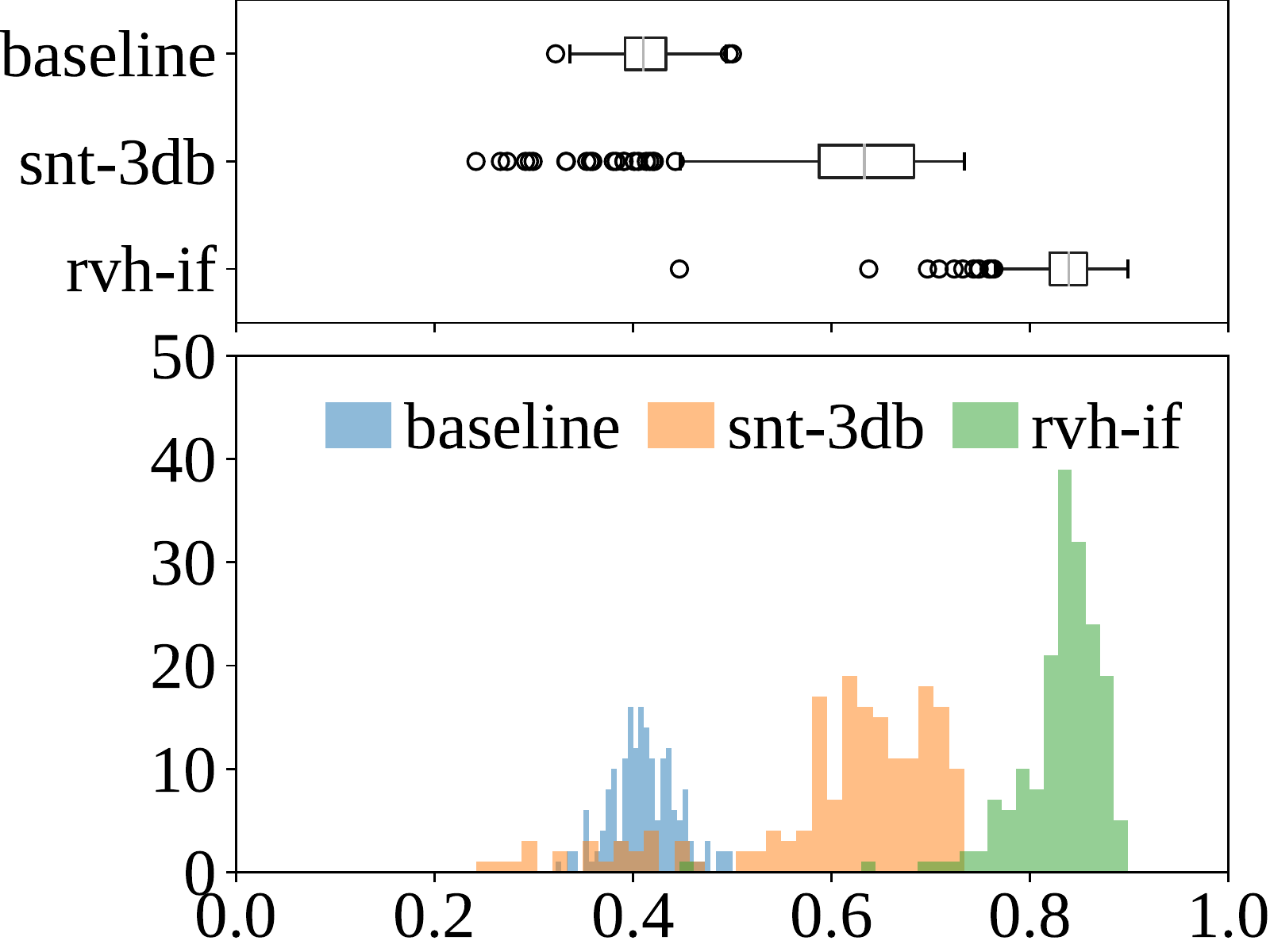}
  \caption{
    Challenge 1 - track 2:
    Boxplots and frequency distributions of shape scores on the test set,
    for the baseline unmodified partial data
    and all submissions, SnT-3DB and RVH-IF.
  }
  \label{fig:score_distribution_1_2}
\end{figure}

Challenge 1 - Track 2 targets the reconstruction of fine details on the hands,
ears and nose, with only a shape ground truth.
The only score evaluated is thus the shape score.
As in Track 1, and as visible in Fig.~\ref{fig:score_distribution_1_2},
RVH-IF is the highest-performing approach
with a score of 83\% compared to
60.7\% for SnT-3DB
and 41.1\% for the baseline.
The distribution of scores is also more concentrated around higher values for
RVH-IF.

\subsection{Challenge 2}
\label{sec:results_2}

Challenge 2 contains four submissions by a single participant
with the RVH-IF method.
As reported in Table~\ref{tab:results}, RVH-IF-2 achieves the highest score
around 75\%, which is more than 30\% over the baseline.
The RVH-IF-1 is the second best-performing approach, a few percent below.

\begin{figure}
  \centering
  \begin{subfigure}[b]{0.32\textwidth}
    \includegraphics[width=\textwidth]{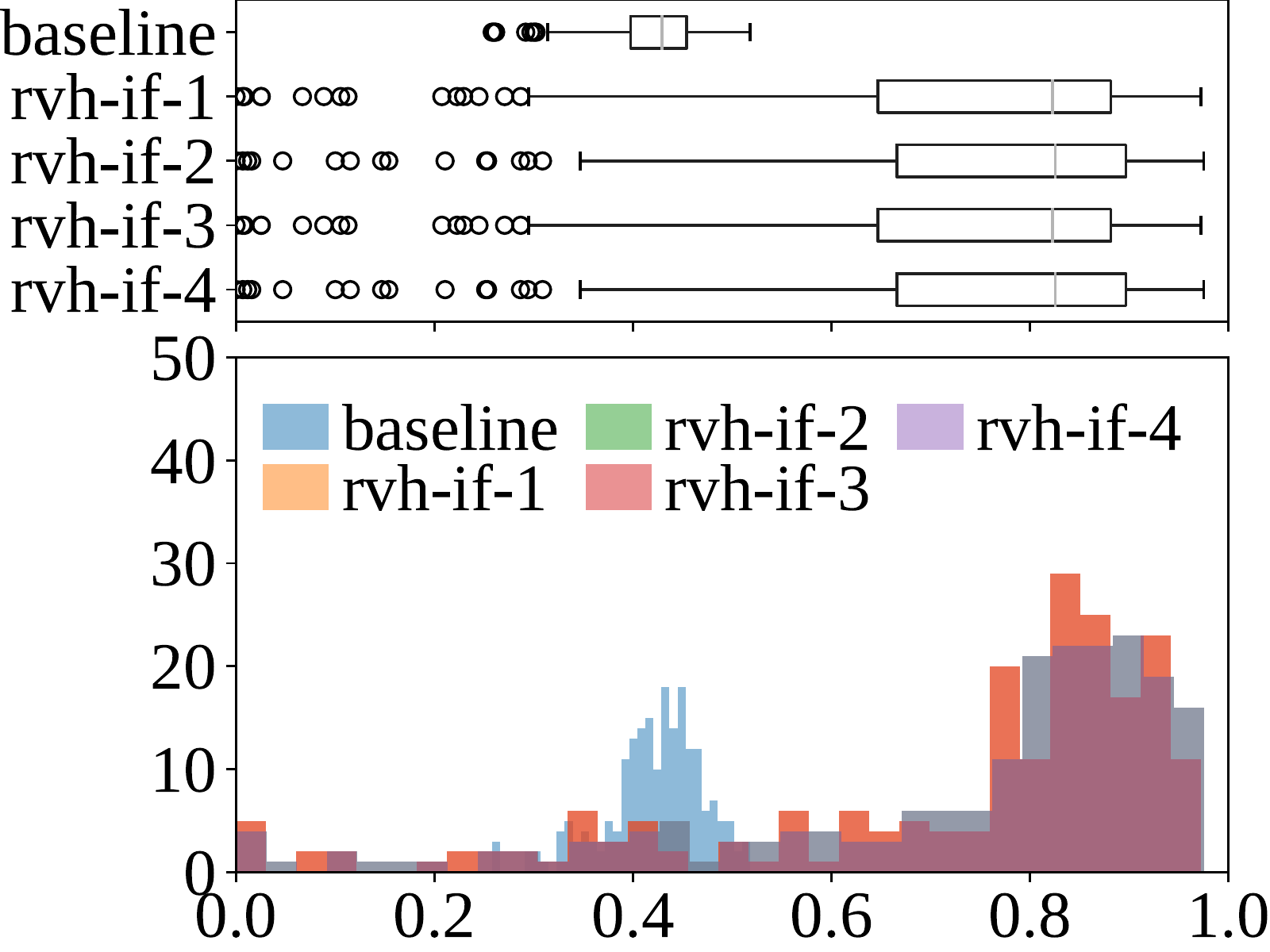}
    \caption{shape score}
  \end{subfigure}
  \begin{subfigure}[b]{0.32\textwidth}
    \includegraphics[width=\textwidth]{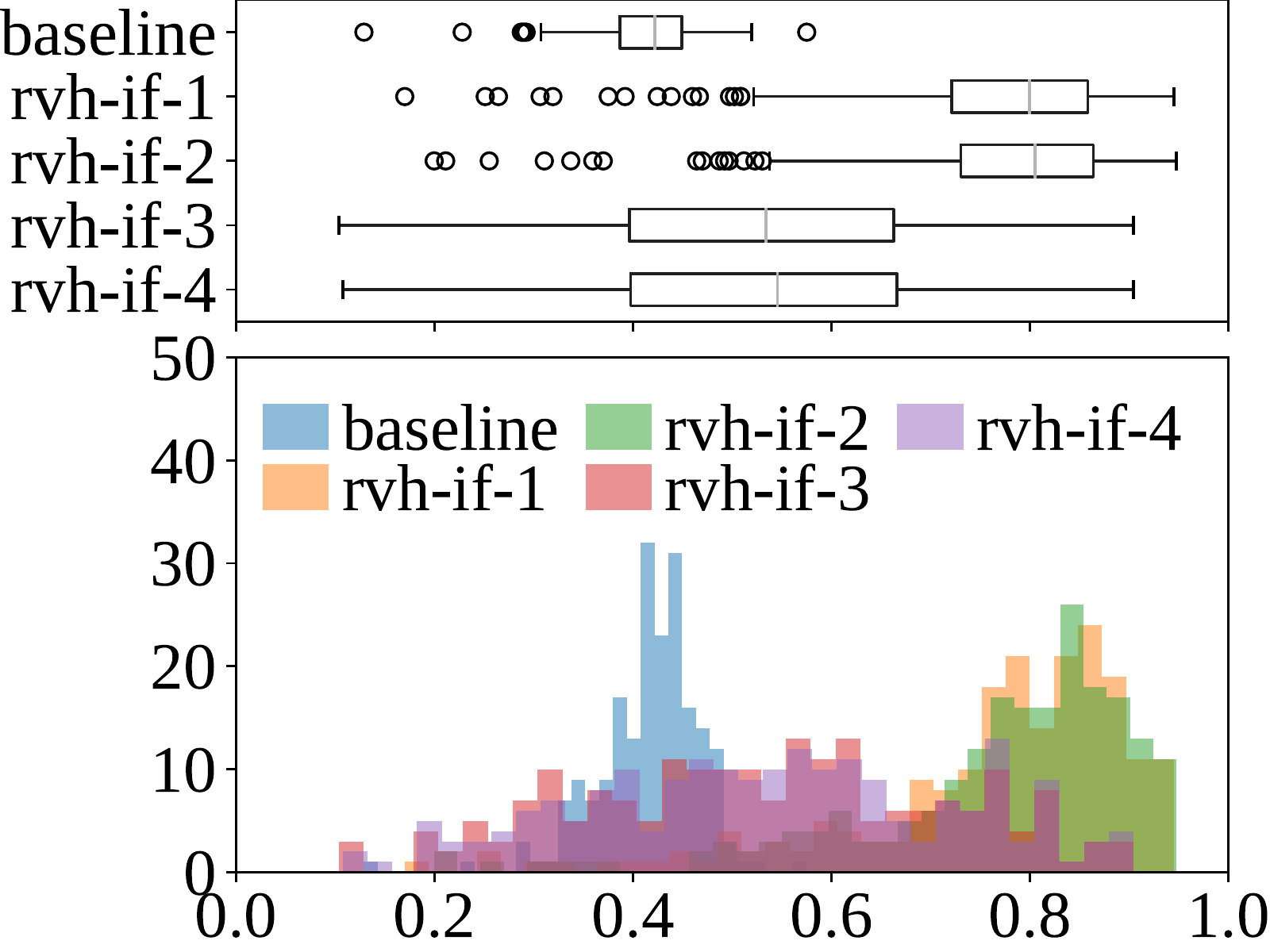}
    \caption{texture score}
  \end{subfigure}
  \begin{subfigure}[b]{0.32\textwidth}
    \includegraphics[width=\textwidth]{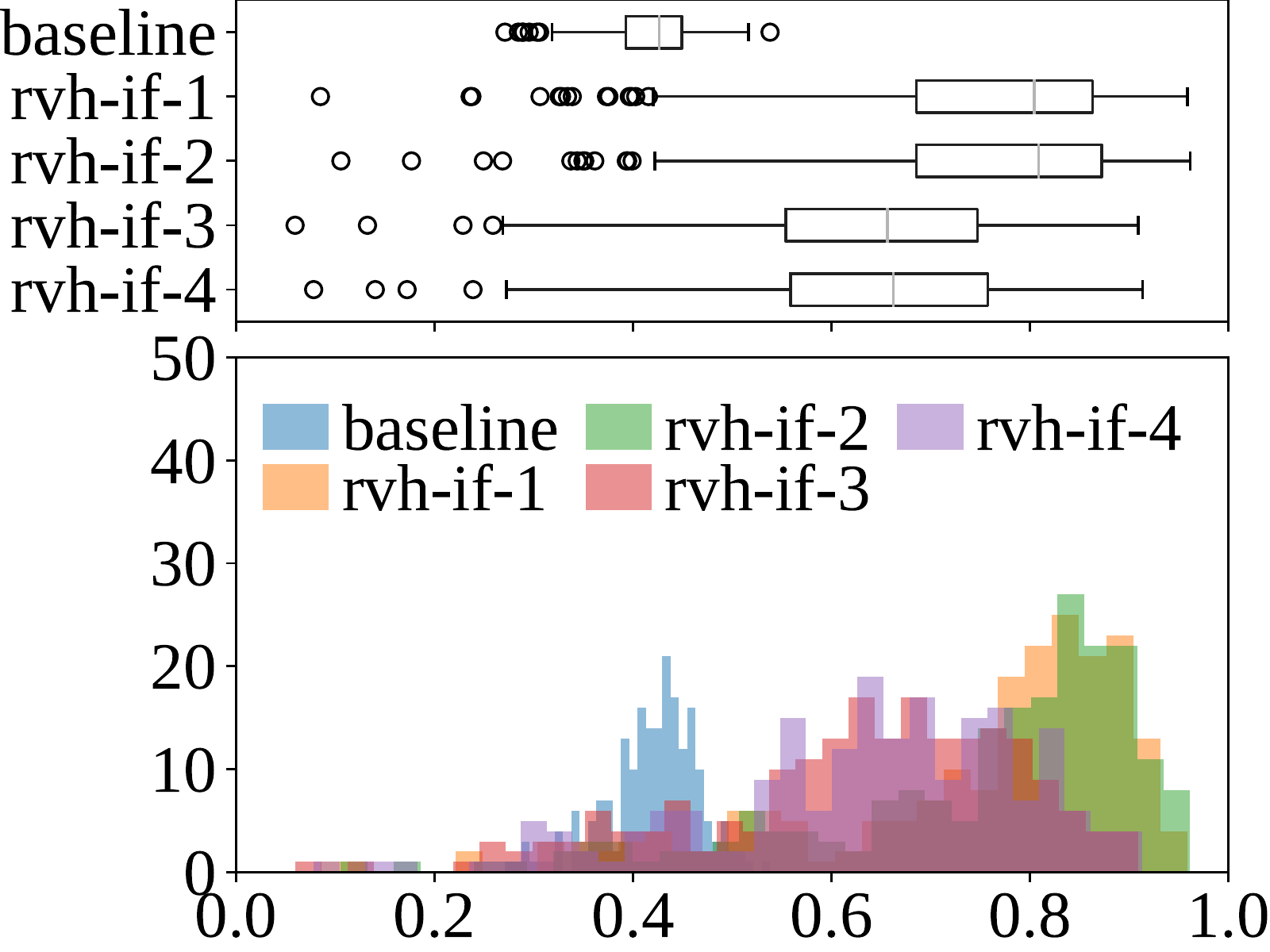}
    \caption{overall score}
  \end{subfigure}
  \caption{
    Challenge 2:
    Boxplots and frequency distributions of scores on the test set
    for the baseline unmodified partial data,
    and all submissions RVH-IF-\{1,4\}.
  }
  \label{fig:score_distribution_2}
\end{figure}

\begin{figure}
  \centering
  \includegraphics[width=.32\textwidth]{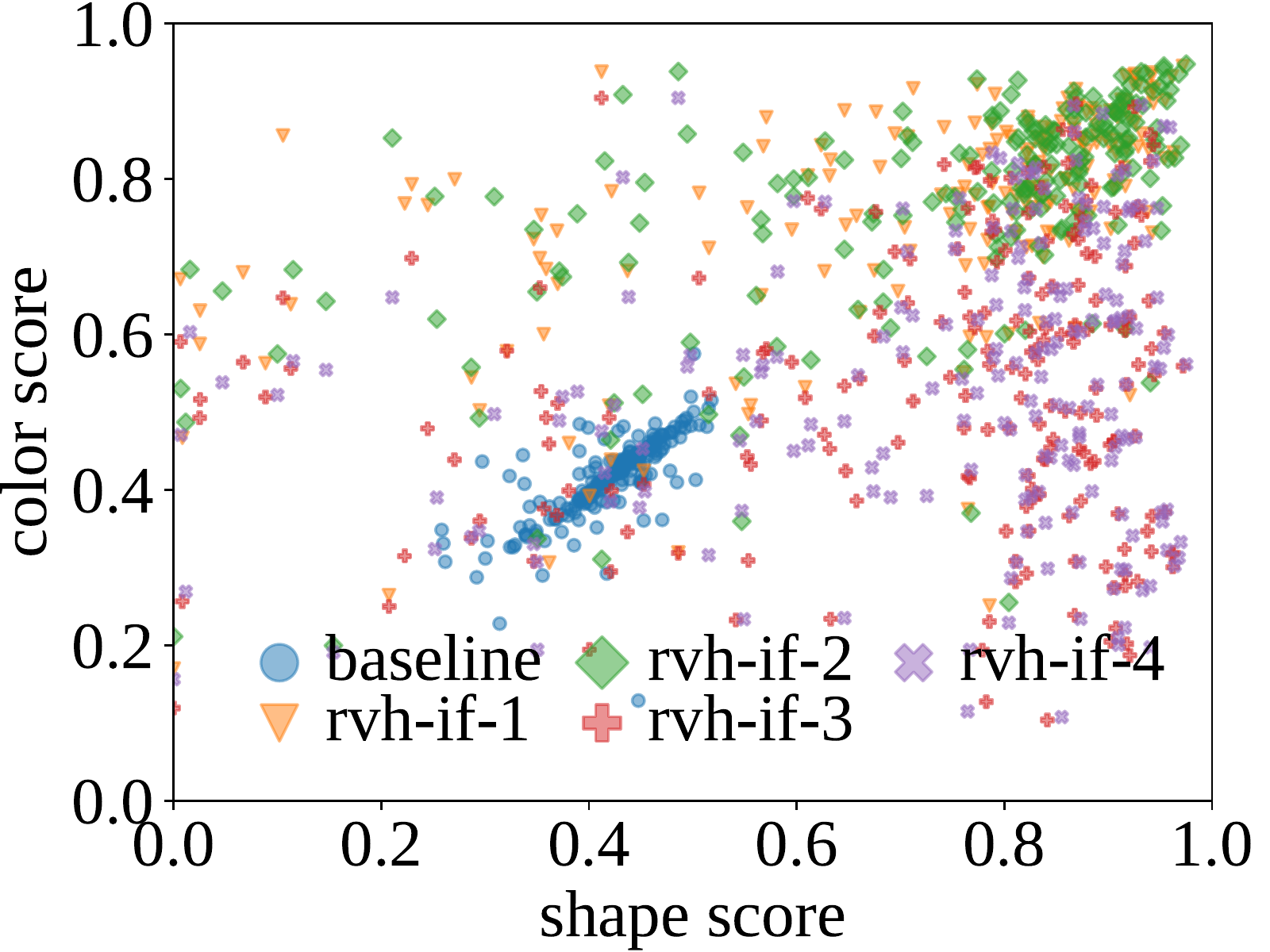}
  \caption{
    Challenge 2:
    Correlation between shape and texture on the test set,
    for the baseline unmodified partial data,
    and all submissions RVH-IF-\{1,4\}.
  }
  \label{fig:correlation_2}
\end{figure}

Fig.~\ref{fig:score_distribution_2} shows the distribution of reconstruction
scores.
All approaches achieve high shape scores.
For the texture, RVH-IF-\{3,4\} perform less well and overlap the baseline,
not improving much on the unmodified partial data.
Overall, the scores are widely spread, even for the baseline.
This reflects the diversity of object classes in 3DObjectTex,
making both the reconstruction and the evaluation tasks challenging.
This is confirmed in the shape-texture correlation in
Fig.~\ref{fig:correlation_2}
where the correlation is less than in Challenge~1.

\begin{figure}
\centering
\begingroup
\setlength{\tabcolsep}{0pt}
\newcommand\widthfactor{.125}
\begin{tabular}{@{}*{8}{c}@{}}
  \includegraphics[width=\widthfactor\textwidth]{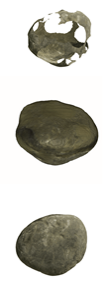}
  & \includegraphics[width=\widthfactor\textwidth]{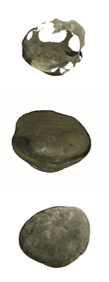}
  & \includegraphics[width=\widthfactor\textwidth]{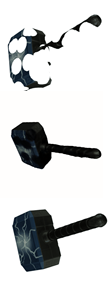}
  & \includegraphics[width=\widthfactor\textwidth]{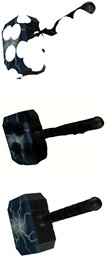}
  & \includegraphics[width=\widthfactor\textwidth]{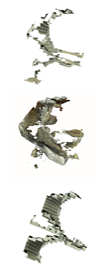}
  & \includegraphics[width=\widthfactor\textwidth]{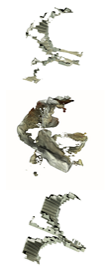}
  & \includegraphics[width=\widthfactor\textwidth]{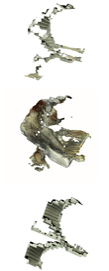}
  & \includegraphics[width=\widthfactor\textwidth]{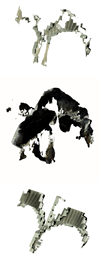}
  \\
  RVH-IF-1 & RVH-IF-2 & RVH-IF-3 & RVH-IF-4
  &
  RVH-IF-1 & RVH-IF-2 & RVH-IF-3 & RVH-IF-4
  \\
  \cmidrule(lr){1-4}
  \cmidrule(lr){5-8}
  \multicolumn{4}{c}{best}
  &
  \multicolumn{4}{c}{worst}
  \\
\end{tabular}
\endgroup

\caption{
  Challenge 2:
  Best and worst reconstruction results for all submissions (RVH-IF-\{1,4\}).
  From top to bottom:
  partial scan,
  reconstruction,
  ground truth.
}
\label{fig:best_worst_2}
\end{figure}

As Fig.~\ref{fig:best_worst_2} shows,
the best reconstructions are of meshes
with a closed surface,
and with both a smooth shape and a relatively uniform texture.
The worst reconstructions are of meshes with complex shapes,
\ie non-convex and with holes, creases, sharp edges, steps, etc.

%% file: conclusion.tex
\section{Conclusion}
\label{sec:discussion}

This paper presented the SHARP 2020 challenge for
\emph{SHApe Recovery from Partial textured 3D scans}
held in its first edition in conjunction with the
16th European Conference on Computer Vision.
The SHARP challenge proposes to foster research and provide a benchmark on 3D shape
and texture completion from partial 3D scan data.
With two aspects, the recovery of human scans and of object scans,
two unique datasets of reference high-quality textured 3D scans were proposed
and released to the scientific community.
Moreover, new specific evaluation metrics were proposed to measure
simultaneously the quality of shape and texture reconstruction,
and the amount of completion.
The results of the participants
show the validity of the proposed metrics,
the challenging nature of the datasets
and highlight the difficulties of the task.
The RVH-IF submission obtained the highest scores in both challenges with
the \emph{Implicit Feature Networks for Texture Completion of 3D Data}.
The variation in clothing seems a major difficulty in Challenge 1, and to a
lesser extent, the pose.
The texture seems more easily reconstructed than the shape, probably due to it
being mostly uniform on clothing.
The reconstruction of fine details was more demanding than the reconstruction of the full body.
Challenge 2 shows that the proposed dataset of generic objects contains a lot
of variation that makes both the shape and the texture challenging to recover.
As in Challenge 1, smooth objects were better reconstructed.

The SHARP 2020 challenge has thus promoted the development of new methods for
3D shape and texture completion.
The addition of the texture aspect on top of the shape aspect makes the
contributions stand out from the current scientific literature.
As the proposed metrics show, there is still room for improvement.
This will be the object of the next editions of SHARP challenges.

%% file: acknowledgements.tex
\section*{Acknowledgements}
We thank Artec3D for sponsoring this challenge with cash prizes
and releasing the data for the 3DObjectTex dataset.
This work and the data collection of 3D human scans for 3DBodyTex.v2
were partly supported by the Luxembourg National Research Fund (FNR) (11806282 and 11643091).
We also gratefully acknowledge the participation, at different times, of all members of the Computer Vision, Imaging and Machine Intelligence (CVI$^2$) Research Group at the SnT, University of Luxembourg, including the moderation of the workshop event by Renato Baptista and the support of Pavel Chernakov in the development of the evaluation software. Finally, we express our appreciation to all the reviewers of the workshop submissions.